\documentclass[]{bytedance_seed}

\usepackage[toc,page,header]{appendix}

\usepackage{minitoc}

\usepackage{amsmath,amssymb,amsthm,graphicx,algorithm,algpseudocode,bbm,mathtools,caption,multirow, makecell, arydshln}
\usepackage[table,xcdraw]{xcolor}

\usepackage{tikz}
\usetikzlibrary{positioning, calc}

\usepackage{xcolor}
\usepackage{listings}
\usepackage{textcomp}
\usepackage[framemethod=TikZ]{mdframed}

\definecolor{codegreen}{rgb}{0,0.6,0}
\definecolor{codegray}{rgb}{0.5,0.5,0.5}
\definecolor{codepurple}{rgb}{0.58,0,0.82}
\definecolor{backcolour}{rgb}{0.97,0.97,0.97}

\lstdefinelanguage{Lean}{
keywords={
theorem, lemma, example, def, meta, inductive, structure, class, instance,
import, open, export, namespace, section, end,
begin, by, have, assume, show, exact, from, calc, let, in,
sorry, simp, rw, refl, apply, intro, cases, induction,
linarith, nlinarith, norm_num, norm_cast, inhabit, set_option,
maxRecDepth, maxHeartbeats,
Prop, Type, Sort, fun, at, if, then, else, omega, interval_cases, contrapose, intros, classical, first, on_goal, repeat', trans, with, abel, use,
repeat', focus, all_goals, any_goals, try, refine, constructor, ext,
rintro, obtain, rcases, aesop, nth_rewrite, nth_rw, rewrite, convert,
congr, field_simp, ring_nf, set, clear, clear_value, nontriviality,
simpa, simp_all, replace, tauto, ring, revert, by_cases, left, erw,
and, or, not, iff, true, false, xor, exists, forall, implies
},
comment=[l]{--},
morecomment=[s]{/-}{-/},
morestring=[b]",
sensitive=true
}

\lstdefinestyle{leanstyle}{
language=Lean,
backgroundcolor=\color{backcolour},
commentstyle=\color{codegreen},
keywordstyle=\color{blue},
stringstyle=\color{codepurple},
basicstyle=\ttfamily\small,
breakatwhitespace=false,
breaklines=true,
captionpos=b,
keepspaces=true,
numbers=none,
numbersep=5pt,
showspaces=false,
showstringspaces=false,
showtabs=false,
tabsize=2,
frame=single,
rulecolor=\color{black!20},
upquote=true,
literate=
{‹}{{$\langle$}}1
{›}{{$\rangle$}}1
{↑}{{$\uparrow$}}1
{ℤ}{{$\mathbb{Z}$}}1
{ℕ}{{$\mathbb{N}$}}1
{ℝ}{{$\mathbb{R}$}}1
{ℚ}{{$\mathbb{Q}$}}1
{ℂ}{{$\mathbb{C}$}}1
{∀}{{$\forall$}}1
{∃}{{$\exists$}}1
{∄}{{$\nexists$}}1
{≤}{{$\leq$}}1
{≥}{{$\geq$}}1
{≠}{{$\neq$}}1
{≈}{{$\approx$}}1
{≡}{{$\equiv$}}1
{≢}{{$\not\equiv$}}1
{→}{{$\rightarrow$}}1
{←}{{$\leftarrow$}}1
{↔}{{$\leftrightarrow$}}1
{⇒}{{$\Rightarrow$}}1
{⇐}{{$\Leftarrow$}}1
{⇔}{{$\Leftrightarrow$}}1
{¬}{{$\neg$}}1
{∧}{{$\land$}}1
{∨}{{$\lor$}}1
{⊕}{{$\oplus$}}1
{⊗}{{$\otimes$}}1
{∈}{{$\in$}}1
{∉}{{$\notin$}}1
{⊆}{{$\subseteq$}}1
{⊂}{{$\subset$}}1
{⊇}{{$\supseteq$}}1
{⊃}{{$\supset$}}1
{∪}{{$\cup$}}1
{∩}{{$\cap$}}1
{∅}{{$\emptyset$}}1
{∖}{{$\setminus$}}1
{×}{{$\times$}}1
{∘}{{$\circ$}}1
{∙}{{$\bullet$}}1
{·}{{$\cdot$}}1
{⊤}{{$\top$}}1
{⊥}{{$\bot$}}1
{⊢}{{$\vdash$}}1
{⊣}{{$\dashv$}}1
{λ}{{$\lambda$}}1
{Λ}{{$\Lambda$}}1
{Π}{{$\Pi$}}1
{Σ}{{$\Sigma$}}1
{∑}{{$\sum$}}1
{∏}{{$\prod$}}1
{∫}{{$\int$}}1
{∂}{{$\partial$}}1
{∇}{{$\nabla$}}1
{∞}{{$\infty$}}1
{α}{{$\alpha$}}1
{β}{{$\beta$}}1
{γ}{{$\gamma$}}1
{δ}{{$\delta$}}1
{ε}{{$\varepsilon$}}1
{ζ}{{$\zeta$}}1
{η}{{$\eta$}}1
{θ}{{$\theta$}}1
{ι}{{$\iota$}}1
{κ}{{$\kappa$}}1
{μ}{{$\mu$}}1
{ν}{{$\nu$}}1
{ξ}{{$\xi$}}1
{π}{{$\pi$}}1
{ρ}{{$\rho$}}1
{σ}{{$\sigma$}}1
{τ}{{$\tau$}}1
{φ}{{$\varphi$}}1
{χ}{{$\chi$}}1
{ψ}{{$\psi$}}1
{ω}{{$\omega$}}1
{Ω}{{$\Omega$}}1
{₀}{{$_0$}}1
{₁}{{$_1$}}1
{₂}{{$_2$}}1
{₃}{{$_3$}}1
{₄}{{$_4$}}1
{₅}{{$_5$}}1
{₆}{{$_6$}}1
{₇}{{$_7$}}1
{₈}{{$_8$}}1
{₉}{{$_9$}}1
{ₐ}{{$_a$}}1
{ₑ}{{$_e$}}1
{ᵢ}{{$_i$}}1
{ⱼ}{{$_j$}}1
{ₖ}{{$_k$}}1
{ₘ}{{$_m$}}1
{ₙ}{{$_n$}}1
{ₒ}{{$_o$}}1
{ₚ}{{$_p$}}1
{ᵣ}{{$_r$}}1
{ₛ}{{$_s$}}1
{ₜ}{{$_t$}}1
{ᵤ}{{$_u$}}1
{ᵥ}{{$_v$}}1
{ₓ}{{$_x$}}1
{⁰}{{$^0$}}1
{¹}{{$^1$}}1
{²}{{$^2$}}1
{³}{{$^3$}}1
{⁴}{{$^4$}}1
{⁵}{{$^5$}}1
{⁶}{{$^6$}}1
{⁷}{{$^7$}}1
{⁸}{{$^8$}}1
{⁹}{{$^9$}}1
{ⁿ}{{$^n$}}1
{ⁱ}{{$^i$}}1
{⁺}{{$^+$}}1
{⁻}{{$^-$}}1
{⁼}{{$^=$}}1
{⁽}{{$^($}}1
{⁾}{{$^)$}}1
{ℓ}{{$\ell$}}1
{ℏ}{{$\hbar$}}1
{ℜ}{{$\Re$}}1
{ℑ}{{$\Im$}}1
{℘}{{$\wp$}}1
{⟨}{{$\langle$}}1
{⟩}{{$\rangle$}}1
{⟦}{{$\llbracket$}}1
{⟧}{{$\rrbracket$}}1
{⌊}{{$\lfloor$}}1
{⌋}{{$\rfloor$}}1
{⌈}{{$\lceil$}}1
{⌉}{{$\rceil$}}1
{‖}{{$|$}}1
{∥}{{$\parallel$}}1
{⊙}{{$\odot$}}1
{⊎}{{$\uplus$}}1
{⊓}{{$\sqcap$}}1
{⊔}{{$\sqcup$}}1
{⊑}{{$\sqsubseteq$}}1
{⊒}{{$\sqsupseteq$}}1
{⊏}{{$\sqsubset$}}1
{⊐}{{$\sqsupset$}}1
{†}{{$\dagger$}}1
{‡}{{$\ddagger$}}1
{★}{{$\star$}}1
{♯}{{$\sharp$}}1
{♭}{{$\flat$}}1
{♮}{{$\natural$}}1
{◇}{{$\Diamond$}}1
{□}{{$\Box$}}1
{△}{{$\triangle$}}1
{▽}{{$\triangledown$}}1
{◁}{{$\triangleleft$}}1
{▷}{{$\triangleright$}}1
{◯}{{$\bigcirc$}}1
{∎}{{$\blacksquare$}}1
{■}{{$\blacksquare$}}1
{□}{{$\square$}}1
{▪}{{$\blacksquare$}}1
{▫}{{$\square$}}1
{✓}{{$\checkmark$}}1
{✗}{{$\times$}}1
{⋮}{{$\vdots$}}1
{⋯}{{$\cdots$}}1
{⋱}{{$\ddots$}}1
{…}{{$\ldots$}}1
{‥}{{$\cdots$}}1
{⋅}{{$\cdot$}}1
{∴}{{$\therefore$}}1
{∵}{{$\because$}}1
{∼}{{$\sim$}}1
{≃}{{$\simeq$}}1
{≅}{{$\cong$}}1
{≈}{{$\approx$}}1
{≍}{{$\asymp$}}1
{≐}{{$\doteq$}}1
{≔}{{$\coloneqq$}}1
{≕}{{$\eqqcolon$}}1
{≙}{{$\circeq$}}1
{≜}{{$\triangleq$}}1
{≝}{{$=^{\text{def}}$}}1
{≟}{{$\stackrel{?}{=}$}}1
{≪}{{$\ll$}}1
{≫}{{$\gg$}}1
{∣}{{$\mid$}}1
{≺}{{$\prec$}}1
{≻}{{$\succ$}}1
{≼}{{$\preceq$}}1
{≽}{{$\succeq$}}1
{⊊}{{$\subsetneq$}}1
{⊋}{{$\supsetneq$}}1
}

\newmdenv[
linecolor=black!70,
linewidth=1.5pt,
roundcorner=5pt,
backgroundcolor=gray!5,
frametitlefont=\bfseries,
frametitlerule=true,
frametitlebackgroundcolor=gray!20,
innertopmargin=10pt,
innerbottommargin=10pt,
skipabove=1.5em,
skipbelow=0.5em
]{casebox}

\newcommand{\prooftype}[1]{\vspace{1em}\noindent\textbf{#1}\par\vspace{0.5em}}

\title{Scaling up Multi-Turn Off-Policy RL and Multi-Agent Tree Search for LLM Step-Provers}

\author[1,\dagger,\P]{Ran Xin}
\author[2,*,\P]{Zeyu Zheng}
\author[3,*,\P]{Yanchen Nie}
\author[3]{Kun Yuan}
\author[1,\dagger]{Xia Xiao}

\affiliation[1]{ByteDance Seed}
\affiliation[2]{Carnegie Mellon University}
\affiliation[3]{Peking University}

\contribution[\P]{Equal contribution}
\contribution[*]{Work done at ByteDance Seed}
\contribution[\dagger]{Corresponding authors}

\abstract{
The integration of Large Language Models (LLMs) with automated theorem proving has shown immense promise, yet is constrained by challenges in scaling up both training-time reinforcement learning (RL) and inference-time compute. This paper introduces \texttt{BFS-Prover-V2}, an open-source step-level theorem proving system designed to address this dual scaling problem. We present two primary innovations. The first is a novel multi-turn off-policy RL framework for continually improving the performance of the LLM step-prover at training time. This framework, inspired by the principles of AlphaZero, utilizes a multi-stage expert iteration pipeline featuring adaptive tactic-level data filtering and periodic retraining to surmount the performance plateaus that typically curtail long-term RL in LLM-based agents. The second innovation is a planner-enhanced multi-agent system that scales reasoning capabilities at inference time. This architecture employs a general reasoning model as a high-level planner to iteratively decompose complex theorems into a sequence of simpler subgoals. This hierarchical approach substantially reduces the search space, enabling a team of parallel prover agents to collaborate efficiently by leveraging a shared proof cache. We demonstrate that this dual approach to scaling yields state-of-the-art results on established formal mathematics benchmarks. \texttt{BFS-Prover-V2} achieves 95.08\% and 41.4\% on the miniF2F and ProofNet test sets respectively. While demonstrated in the domain of formal mathematics, the RL and inference techniques presented in this work are of broader interest and may be applied to other domains requiring long-horizon multi-turn reasoning and complex search. Our models and code have been open-sourced at \url{https://github.com/ByteDance-Seed/BFS-Prover-V2}.
}

\date{September 9, 2025}
\correspondence{\{ran.xin, x.xiaxiao\}@bytedance.com}
\checkdata[Project Page]{\url{https://bfs-prover.github.io/V2/}}

\begin{document}
\maketitle

\section{Introduction}
Automated Theorem Proving (ATP), a subfield of mathematical logic and automated reasoning, represents one of the foundational ambitions of computer science~\citep{history_deduction}. The contemporary landscape of formal mathematics is increasingly dominated by interactive theorem provers (ITPs) or proof assistants. These systems, such as Coq, Isabelle, and Lean, require a human user to guide the proof process, but they automate significant deductive tasks and, most importantly, provide a machine-checkable guarantee of correctness~\cite{atp_history}. Among these, the Lean4 programming language~\citep{Lean4} has emerged as a particularly vibrant ecosystem. A key factor in its success is Mathlib~\citep{mathlib}, a vast and comprehensive, community-driven library of formalized mathematics. Spanning over a million lines of code, mathlib covers extensive areas of algebra, analysis, topology, and more, providing a rich foundation for both advanced mathematical research and the development of verified systems. 

The rise of Lean4 has coincided with the explosion in the capabilities of LLMs~\citep{openai2023gpt4,gemini2.5,seed1.5-thinking}, opening a new frontier in neuro-symbolic AI systems. The goal here is to integrate the intuitive yet powerful generation and search capabilities of LLMs with the absolute logical verification of formal systems. This research direction centers on a key feedback loop: an LLM proposes intuitive proof steps, the Lean compiler provides rigorous verification, and RL~\cite{sutton2018reinforcement} uses that verification to continuously improve the LLM's reasoning abilities~\citep{ai4formal,deepseek_prover_v1,polu2022formal,han2021proofartifact,lample2022hypertree}. 

\subsection{A Duality of Scaling Challenges in LLM Provers and Reasoning Agents}
The development of high-performance LLM-based provers, or any other reasoning agents, is contingent upon solving two fundamental and deeply interconnected scaling challenges. 

\textbf{Training-time scaling.} This refers to the techniques required to continuously enhance a model's foundational capabilities and tactical intuitions via training. A common and significant obstacle in applying RL to LLMs is the phenomenon of performance plateaus: after an initial phase of rapid improvement, models often stagnate, with their capabilities ceasing to grow despite continued training~\citep{Prorl,kimi-RL,dapo,vapo,ds-r1,seed1.5-thinking,deepseek_prover_v1,deepseek-proer-v1.5}. Overcoming this limitation requires carefully designed algorithms that can sustain learning over extended periods, enabling the model to transition from mastering simple problems to tackling increasingly complex theorems. 

\textbf{Inference-time scaling.} This addresses the method of using a trained model to solve previously unseen theorems. Real-world mathematical problems often require deep, multi-step reasoning, the formulation of intermediate lemmas, and the exploration of an exponentially large search space of possible tactics. A powerful base model, while necessary, is not sufficient. Without an effective search strategy, even a competent model will be overwhelmed by the combinatorial search complexity. The challenge, therefore, is to design an inference architecture that can efficiently decompose complex problems into simpler parts, and strategically allocate computational resources to the most promising avenues of exploration~\citep{prover-agent,delta-prover,seed-prover,dsp,dsp+}. 

\subsection{Our Contributions}
This paper presents \texttt{BFS-Prover-V2}, a comprehensive training and inference system for neural theorem proving in Lean4 that introduces novel solutions to the above scaling challenges. The primary contributions of this work are as follows: 

\textbf{Novel RL Scaling Techniques at Training:} We develop a distillation-free multi-stage expert-iteration framework~\citep{alphazero,expert-iteration}, a form of off-policy RL, tailored for the domain of formal theorem proving.
To sustain learning and overcome performance plateaus, we introduce a suite of specialized techniques within the RL pipeline. These include an adaptive, perplexity-based data filtering strategy at the tactic level, which creates an automated curriculum for the agent, and a periodic retraining mechanism that acts as a ``soft reset'' to escape local optima in the model parameter space and increase model scaling potential.

\textbf{A Planner-Enhanced Multi-Agent Tree Search System at Inference:} For inference-time scaling, we introduce a hierarchical reasoning architecture. A general-purpose reasoning model, termed the Planner, iteratively decomposes complex theorems/goals into a sequence of more manageable subgoals. These subgoals are then tackled by a group of parallel prover agents that share a common subgoal cache, dramatically decreasing search complexity of the system and enabling it to solve problems that are intractable for a monolithic prover.

\textbf{State-of-the-Art Empirical Results:} We validate the effectiveness and generalizability of our dual scaling approach on established benchmarks. \texttt{BFS-Prover-V2} achieves 95.08\% on miniF2F test set, largely surpassing previous step-provers~\citep{intern-prover-v2.5,bfs-prover-v1} and performing on par with best whole-proof models~\citep{ds-prover-v2,goedel-prover-v2,kimina-prover}. On ProofNet test, it achieves 41.4\%, setting a new state-of-the-art, showing robust generalization across distributions.

\section{The BFS-Prover-V2 System}
This section details the two core components of \texttt{BFS-Prover-V2}: (i) a training pipeline, grounded in a Markov Decision Process (MDP)~\cite{mdp} and scaled via adaptive filtering and periodic retraining; and (ii) an inference engine, which uses a planner-enhanced multi-agent search for hierarchical reasoning. These components build upon the foundation of \texttt{BFS-Prover-V1}~\cite{bfs-prover-v1} to specifically address the dual challenges of scaling at both training and inference time. We provide visual overviews of these components in Fig.~\ref{fig:expert_iteration_pipeline} and~\ref{fig:proof_flowchart}, with practical implementation parameters detailed in Section~\ref{sec:prac}.

\subsection{A Step-Level Formulation: Theorem Proving as a Markov Decision Process}
We formulate proof search in Lean4 tactic mode as a multi-turn interaction between an agent and an~environment, modeled as a MDP. In this formulation, the LLM prover acts as the agent, and the Lean compiler, with its current tactic state, serves as the environment. This approach captures the sequential, stateful nature of constructing a formal proof one step at a time~\cite{Lean4,yang2024leandojo}. The components of our MDP are defined as follows:

\begin{itemize}
    \item \textbf{State ($S$):} A state is the current tactic state as given by the Lean compiler. This includes the hypotheses, i.e., known facts, and the target goals to be proven.
    
    \item \textbf{Action ($A$):} The LLM acts as the policy of the MDP, taking the current state $S$ as input to generate a tactic string, which is the action $A$. A tactic is a command that instructs the Lean compiler to perform a deductive step, such as applying a theorem, rewriting an expression, or breaking the goal into cases.
    
    \item \textbf{Transition ($P(S' \mid S, A)$):} The transition function is deterministically executed by the Lean compiler itself. When the agent submits a tactic (action $A$) in a given state $S$, the compiler attempts to apply it. If the tactic is valid and applicable, the compiler transitions to a new state $S'$. If the tactic is invalid or fails, the compiler returns an error message.
    
    \item \textbf{Reward ($R$):} A reward of $+1$ is given for every state-action pair $(S, A)$ that lies on a successful proof path. All other $(S, A)$ pairs that do not contribute to a final proof receive a reward of 0.
\end{itemize}

This step-level, interactive formulation stands in sharp contrast to whole proof generation models~\citep{ds-prover-v2,goedel-prover-v2,kimina-prover}, which treat theorem proving as a one-shot, code generation task from a theorem statement to a full proof script. While simpler, the existing whole-proof approaches lack the ability to react to the intermediate states of a proof and cannot be easily integrated into the interactive workflow of a human mathematician~\citep{llm-step, lean-copilet}. Our MDP-based approach, by design, trains an agent that functions as a genuine Lean copilot, suggesting the next logical tactic at any point in the proof process~\cite{yang2024leandojo}. 

\subsection{Expert Iteration with Best-First Tree Search}

The core training loop of \texttt{BFS-Prover-V2} is an expert iteration pipeline, which may be viewed as a variant of the AlphaZero algorithm~\cite{expert-iteration,alphazero}. This approach enables the system to learn and improve its theorem-proving capabilities from its own experience~\cite{era_of_experience}. The process, illustrated in the inner loop of Fig.~\ref{fig:expert_iteration_pipeline}, includes two major alternating phases: proof generation and model refinement.

\textbf{Phase 1: Proof Generation:} In this phase, the current best version of the LLM prover, referred to as the expert, is tasked with solving a large corpus of mathematical problems. For this work, we autoformalized approximately 3 million problems~\citep{lean-github,lean-workbook,mathlib,numinamath,goedel-prover-v1,bfs-prover-v1} to serve as the training ground. The expert model is coupled with the best-first tree search (BFS) algorithm used in \texttt{BFS-Prover-V1}~\cite{bfs-prover-v1} to explore the vast space of possible proof paths. This combination of a neural policy and systematic search allows the system to find proofs for problems that would be intractable for the model alone. Each successful proof found during this phase constitutes a trajectory of (state, tactic) pairs. Across a single round of expert iteration, the system performs approximately $10^7$ tree searches, generating a massive synthetic dataset.

\textbf{Phase 2: Model Refinement:} The experience data generated in the first phase is then used to improve the LLM prover. In particular, the state-tactic pairs from the successful proof trajectories are used to update the model's parameters. The updated model then becomes the new ``expert'' for the next round of iteration.

\subsection{Scaling up Training: Multi-Stage Expert Iteration}
\label{sec:train}
\begin{figure*}[htbp]
    \centering
    \includegraphics[width=\textwidth]{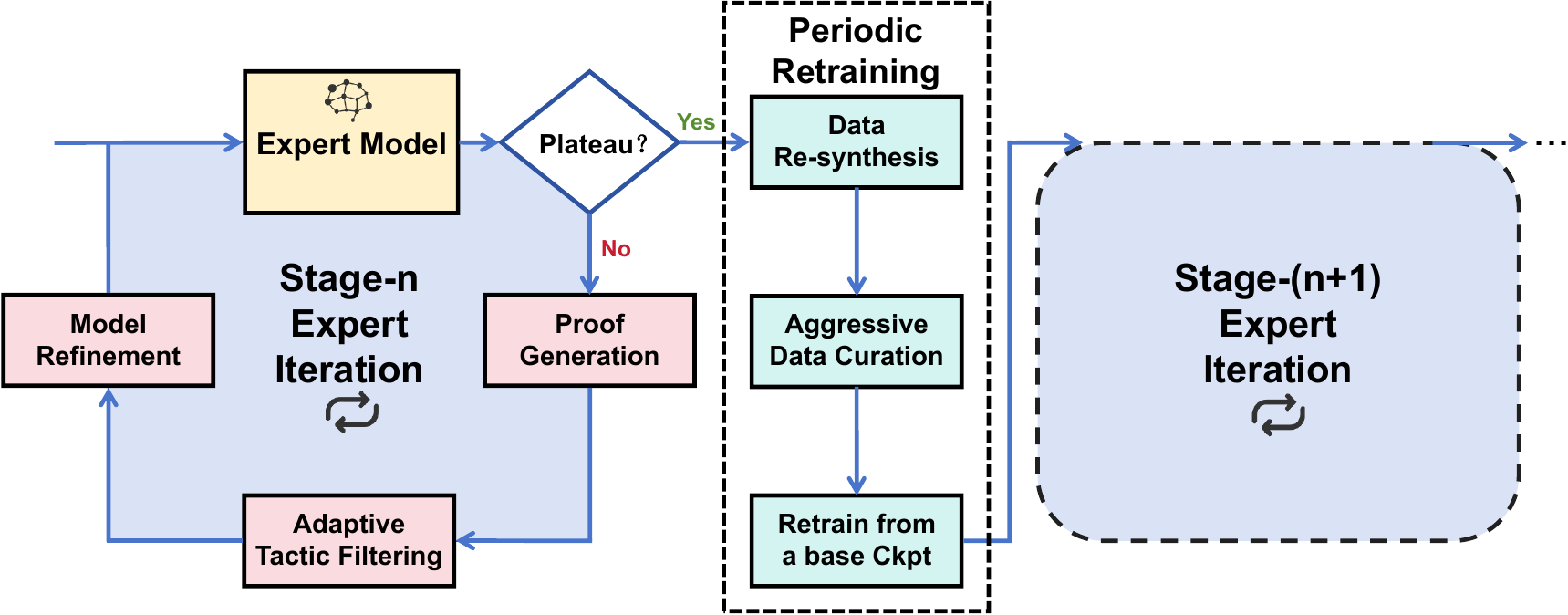} 
    \caption{
        Overview of the training-time scaling up architecture. 
        The process begins with a current expert model. The system then evaluates the model's performance to check for a plateau, which determines the subsequent path. If performance is improving, the model enters an inner \textbf{expert iteration loop} that involves generating new proofs, applying \textit{Adaptive Tactic Filtering}, and refining the model. Conversely, if performance has plateaued, the system triggers the outer \textbf{retraining loop}, which consists of \textit{Data Re-synthesis}, \textit{Aggressive Data Curation}, and \textit{retraining the model from a base checkpoint}. Upon completion, this retraining loop yields a new, improved expert model, which then serves as the starting point for the next cycle of evaluation and iteration.
    }
    \label{fig:expert_iteration_pipeline}
\end{figure*}

A central challenge in scaling the expert iteration pipeline or RL in general is managing the vast quantity and variable quality of the self-generated data. Naively training on every successful tactic discovered during proof generation quickly leads to diminishing returns, performance stagnation, and mode collapse~\cite{Prorl,sutton2018reinforcement,bfs-prover-v1}. To sustain improvement over many iterations, we introduce two key algorithmic innovations: a dynamic, fine-grained data filtering strategy and a periodic full-model retraining process. These techniques work in concert to form a sophisticated, automated curriculum that continuously improves the agent capability in a long horizon. The overall architecture of this pipeline is illustrated in Fig.~\ref{fig:expert_iteration_pipeline}, and we detail each of these innovations in the following subsections.

\subsubsection{Adaptive Tactic Filtering: Learning from the ``Just Right'' Data}
Instead of relying on coarse, problem-level filtering~\cite{dapo,kimi-RL}, which often uses static metrics of difficulty, we adopt a more dynamic and fine-grained approach at the tactic level. This strategy is guided by the empirical observation that the perplexity (negative log-probability) of tactics generated by the LLM follows a roughly Gaussian distribution. The distribution, shown in Fig.~\ref{fig:tactic_filter}, can be divided into three distinct regions, each with different implications for learning:

\begin{figure*}[htbp]
    \centering
    \includegraphics[width=1.0\textwidth]{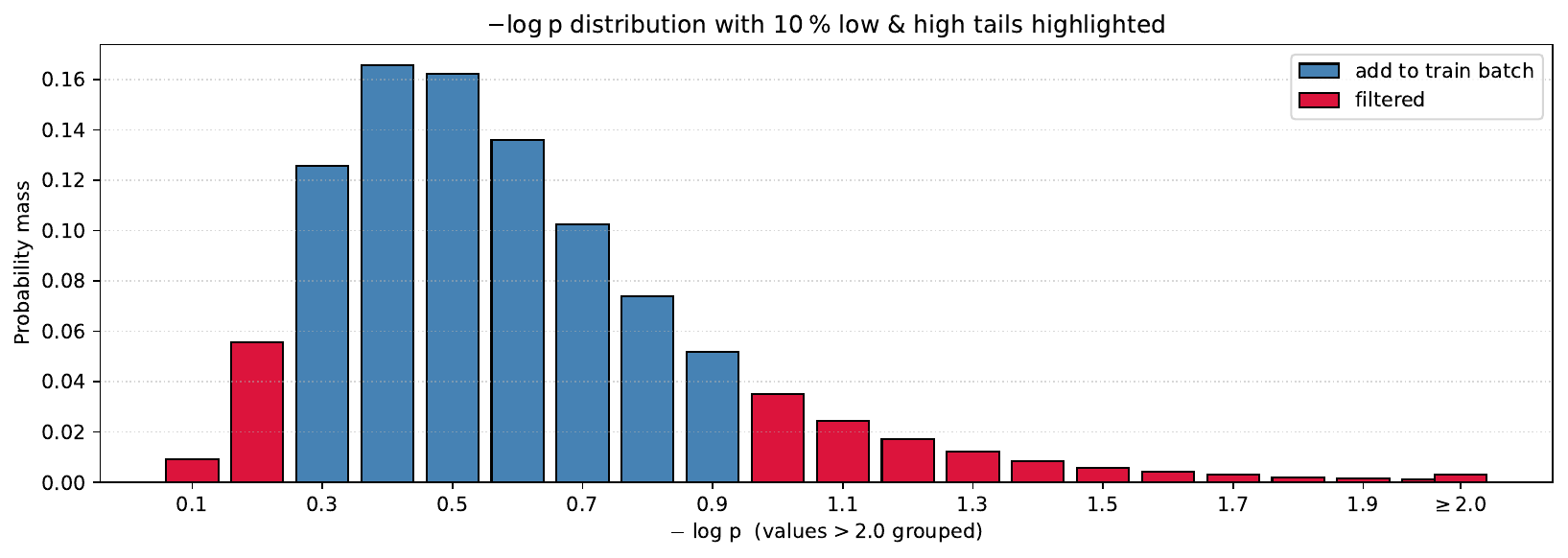}
    \caption{Tactic-Level Data Filtering Based on the Perplexity Distribution. This histogram shows the probability distribution of tactic perplexity (represented as negative log-probability) from a single round of expert iteration. We filter out the low- and high-perplexity tails, shown in red. The low-perplexity tail represents overly simple tactics the model is already confident in, while the high-perplexity tail often consists of noisy or unnecessarily complex tactics. By training only on the central part of the distribution (blue), we focus the model's learning on challenging yet meaningful examples, which prevents overfitting and encourages a smoother, more stable improvement in reasoning capabilities.}
    \label{fig:tactic_filter}
    \vspace{-0.5cm}
\end{figure*}

\begin{itemize}
     
\item\textbf{The Low-Perplexity Tail:} This region corresponds to tactics for which the model has very high confidence. These are typically simple, ``obvious'' steps, such as basic simplification or applying a clear-cut hypothesis. Including these examples in the training batch offers no new learning signal; it merely reinforces what the model already knows well and can contribute to overfitting and a reduction in exploratory capacity.

\item\textbf{The High-Perplexity Tail:} This region represents tactics that the model finds highly surprising. Our case studies reveal that these are often not instances of brilliant, non-obvious reasoning. Instead, they frequently correspond to noisy or suboptimal choices, such as using a powerful, general-purpose tactic with many unnecessary parameters on a simple problem where a more direct tactic would suffice. These ``fancy'' operations can be detrimental to training, as they may teach the model to generate overly complex or irrelevant tactics, leading to hallucinations and degrading its core reasoning ability.

\item\textbf{The Central Distribution:} This is the ``goldilocks'' zone. The tactics in this region are neither too easy nor too noisy. They represent steps that are challenging for the model but still within its grasp---its zone of proximal development. By selectively training only on the data from this central part of the distribution, we ensure that the model is constantly learning at the edge of its capabilities.

\end{itemize}

This adaptive filtering mechanism functions as a fully automated and continuous form of curriculum learning. It does not rely on any external or predefined metric of difficulty. Instead, it uses the model's own uncertainty (as measured by perplexity) as a live, dynamic signal of what constitutes valuable training data at its current stage of development. This ensures a smooth and stable evolution of the model's internal policy distribution throughout the lengthy RL process, enabling sustained growth in performance.

\subsubsection{Periodic Retraining: A ``Soft Reset'' to Escape Local Optima}

Even with adaptive filtering, after a number of expert iteration updates, the model's performance can still begin to plateau. This occurs because the model's proof-finding ``style'' becomes entrenched. It develops strong biases towards certain types of tactics and proof strategies, effectively getting trapped in a local optimum in the vast space of possible reasoning policies. It becomes very good at solving problems in particular ways, but loses the ability to discover novel approaches required for new and harder classes of problems.

\begin{figure*}[htbp]
    \centering
    \includegraphics[width=\textwidth]{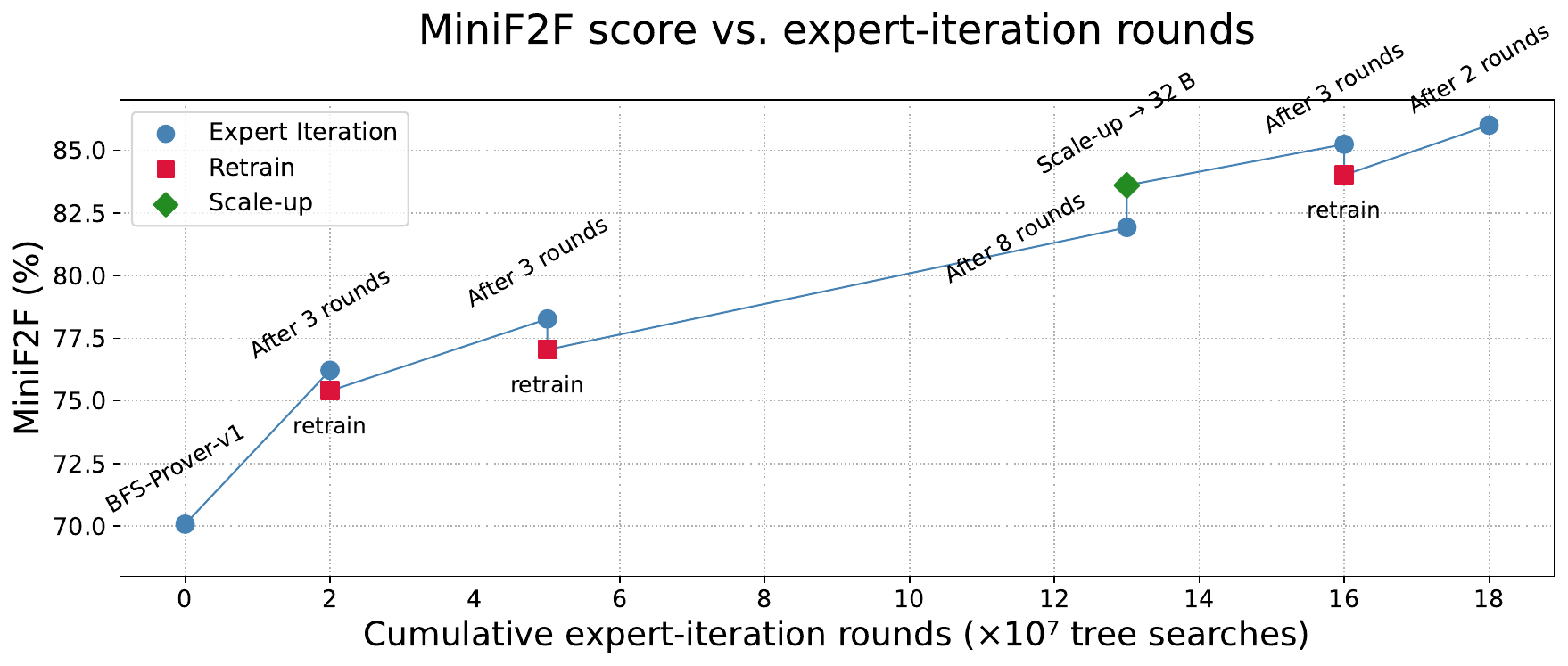}
    \caption{Sustained Performance Improvement through Expert Iteration and Periodic Retraining. This graph plots the prover's performance on miniF2F against the number of expert iteration rounds. Performance steadily increases (blue circles) but eventually begins to plateau as the model settles into a local optimum. To counteract this, we periodically conduct a "soft reset" (red squares). This involves using the current expert model to solve all past problems to generate a cleaner, more efficient dataset, which is then used to retrain the model from a base checkpoint. This procedure allows the model to break out of its local optimum and continue improving, as evidenced by the significant performance jumps following each retraining phase. A model scale-up to 32 billion parameters is also shown (green diamond).}
    \label{fig:exit_progress}
\end{figure*}

To escape local optima and reinvigorate the learning process, we introduce a periodic ``soft reset'' procedure. This constitutes a multi-stage expert-iteration process designed to increase the model's entropy and reset its exploratory potential without losing the competence it has already gained. The procedure is as follows:

\begin{enumerate}
\item \textbf{Re-synthesis and De-noise:} 
The current best-performing prover is used to re-solve the entire corpus of problems it has encountered in all past iterations. Because the prover is now significantly more capable than it was in earlier rounds, it often finds proofs that are shorter, more direct, and more elegant. This step effectively uses the expert model to de-noise and improve upon its own past work, filtering out the redundant or circuitous steps that were present in the initial, less-informed proofs.

\item \textbf{Aggressive Data Curation:} The new, higher-quality proofs generated in the data re-synthesis phase are then subjected to an aggressive version of the tactic-level perplexity filtering described above. A much larger portion of the data is discarded, retaining only the most crucial and informative tactic steps.

\item \textbf{Retrain from a base Checkpoint:} The existing training data is completely replaced by this new, highly curated, and compact dataset. A fresh model instance is then initialized from a general pre-trained checkpoint and trained from scratch on this refined data.
\end{enumerate}

The resulting model, as illustrated in Fig.~\ref{fig:exit_progress}, initially exhibits a temporary drop in performance on the benchmark. This is expected, as it has been trained on a smaller, more focused dataset and has ``forgotten'' some of its previous stylistic biases. However, this new model possesses a significantly higher exploratory potential. When it is reintroduced into the expert iteration loop, its increased capacity for exploration allows it to discover novel proof strategies that were inaccessible to the previous, over-specialized model. Consequently, its performance rapidly recovers and then surpasses the previous peak, establishing a new, higher performance ceiling. This periodic retraining, marked by the ``retrain'' events in Fig.~\ref{fig:exit_progress}, is a critical mechanism for ensuring long-term, monotonic improvement across dozens of training rounds.

\subsection{Scaling up Inference: Planner-Enhanced Multi-Agent Search}
\label{sec:inference}

While the reinforcement learning pipeline scales the LLM's intrinsic competence, solving genuinely difficult mathematical theorems requires a sophisticated inference-time strategy to manage the immense search space. Many complex proofs in mathematics are not found through a linear sequence of simple deductions but rather through a hierarchical process of identifying and proving crucial intermediate results, or lemmas. 

\subsubsection{Planner-Prover Paradigm for Hierarchical Reasoning}

\begin{figure*}[!htbp] 
    \centering 
    \includegraphics[width=1.0\textwidth]{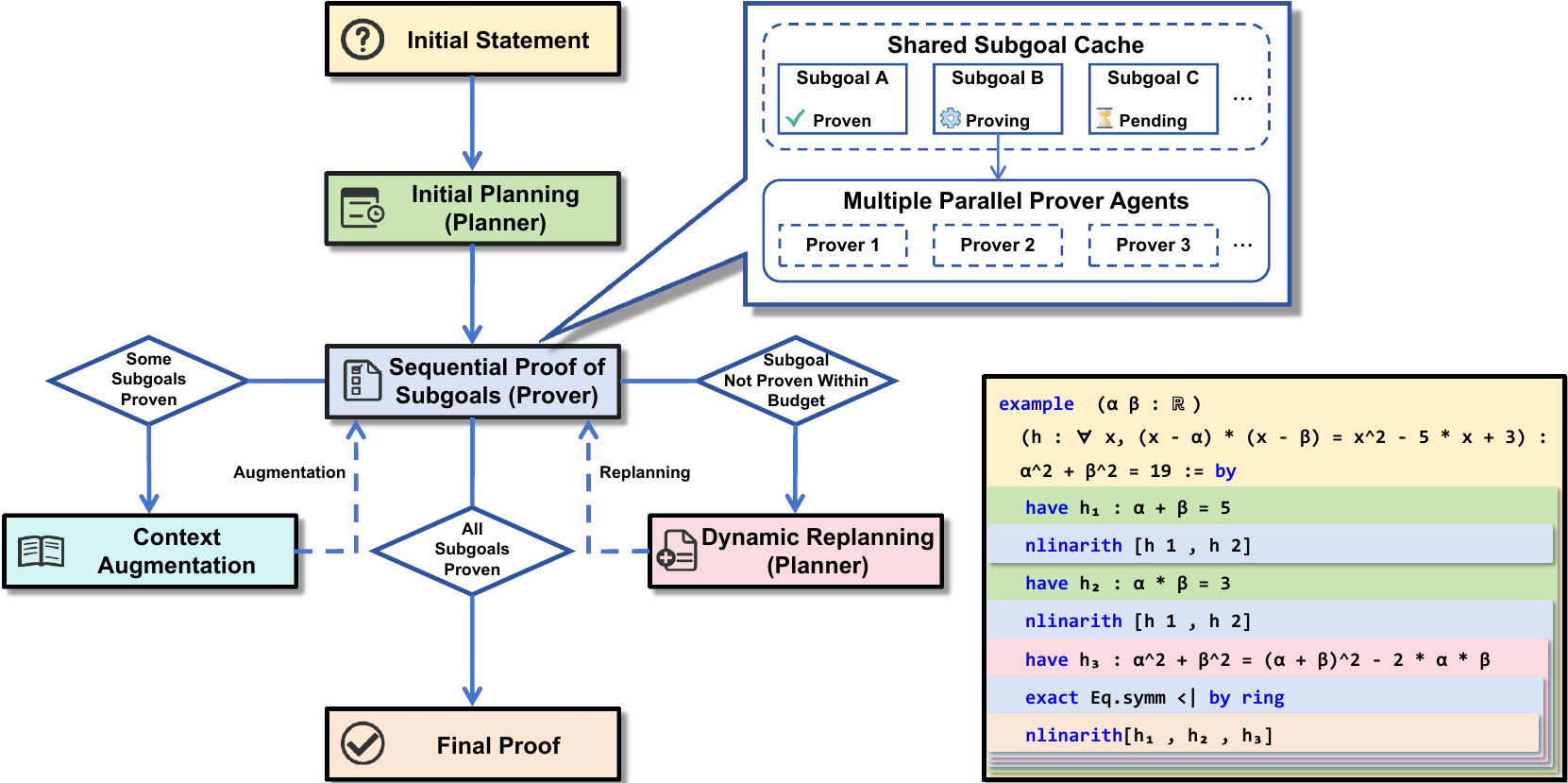} 
    \caption{Overview of the planner-enhanced multi-agent tree search architecture. The \textbf{Planner} agent decomposes the main theorem into a sequence of simpler subgoals, which are managed in a \textbf{Shared Subgoal Cache} and solved in parallel by multiple \textbf{Prover} agents. Successfully proven subgoals augment the main proof's context, while failures can trigger a \textbf{Dynamic Replanning} loop. The inset provides a toy example, demonstrating how proving intermediate lemmas ($\mathrm{h}_1$, $\mathrm{h}_2$, $\mathrm{h}_3$) facilitates the proof of the final goal.}
    \label{fig:proof_flowchart}
\end{figure*}

We introduce a hierarchical inference architecture, shown in Fig.~\ref{fig:proof_flowchart}, that divides the labor of theorem proving between two distinct LLM agents: a high-level planner and a low-level prover.

\textbf{Planner:} This is a general-purpose reasoning LLM tasked with strategic decomposition. Given the current theorem statement and proof progress, its role is not to generate a specific tactic but to propose a high-level plan that includes a series of intermediate subgoals. These subgoals are lemmas that, if proven, would simplify the main proof. By formulating these subgoals, the Planner effectively transforms a single, monolithic, and potentially intractable search problem into a structured sequence of smaller, more manageable ones. This decomposition substantially reduces the dimensionality of the search space that the Prover must explore.

\textbf{Prover:} This is the specialized LLM tactic generator trained via our multi-stage expert iteration pipeline described in Section~\ref{sec:train}. It receives one subgoal at a time from the Planner and uses its learned policy, in conjunction with BFS~\citep{bfs-prover-v1}, to find a formal proof for that specific subgoal. 

This division of labor mirrors the cognitive workflow of a human mathematician, who might first sketch out the high-level structure of a proof by identifying key lemmas (the Planner's role) and then proceed to fill in the detailed, step-by-step deductions for each lemma (the Prover's role). This hierarchical structure is a powerful architectural pattern for tackling complex reasoning tasks.

\subsubsection{Operational Mechanics of Planner-Guided Search}

As shown in Fig.~\ref{fig:proof_flowchart}, the interaction between the Planner and the Prover system unfolds in a dynamic loop, allowing the plan to be revised based on the progress of the proof:

\begin{enumerate}
\item \textbf{Initial Planning:} At the start of a proof attempt, the Planner is queried with the main theorem statement. It returns a list of proposed subgoals, formatted as Lean \texttt{have} statements.

\item \textbf{Sequential Proof of Subgoals:} The Prover system addresses the subgoals one by one. It takes the first subgoal in the queue and initiates a tree search to find its proof.

\item \textbf{Context Augmentation:} Once a subgoal is successfully proven, its statement is ``implanted'' into the main theorem's context. From that point on, the proven subgoal is treated as a known fact, equivalent to a hypothesis, which can be used in the proofs of all subsequent subgoals and the main theorem itself.

\item \textbf{Dynamic Replanning:} If the Prover system fails to find a proof for a subgoal within a given computational budget (i.e., it gets ``stuck''), the process does not terminate. Instead, the Planner is re-queried. This time, however, the input to the Planner now includes all subgoals that were successfully proven before the system got stuck, in addition to the original theorem statement. Taking this new context into account, the Planner generates a revised plan that refines the original proof strategy, often by decomposing the stuck subgoal into a more granular sequence of intermediate steps.
\end{enumerate}

This dynamic and iterative loop between planning and proving makes the \texttt{BFS-Prover-V2} system resilient to getting stuck, effectively scaling its inference-time capabilities to tackle complex theorems that would be intractable for a monolithic tree search.

\subsubsection{Multi-Agent Collaboration via Focused Parallelism and Shared Subgoal Cache}

To accelerate the proof search process and reduce wall-clock time, the Planner-Prover architecture is implemented within a multi-agent framework. Instead of relying on a single Prover agent, we deploy multiple parallel Prover instances that collaborate on the plan generated by the Planner. This collaboration is orchestrated by two key principles: a strategy of focused parallelism and the use of a shared subgoal cache.

\textbf{Focused parallelism:} Rather than assigning different subgoals to different prover agents, our system concentrates the full computational power of all Prover instances on one subgoal at a time. This approach is designed to overcome difficult reasoning bottlenecks that might be intractable for a single agent working alone. Furthermore, this sequential method ensures computational efficiency by preventing agents from wasting resources on later subgoals that would be rendered invalid if an earlier step fails and triggers a replan.

\textbf{Shared Subgoal Cache:} This cache is the central communication and state-tracking mechanism, shared across all parallel Prover instances. It performs several critical functions:
\begin{itemize}

\item It stores the full sequence of subgoals generated by the Planner.

\item It tracks the real-time status of each subgoal (e.g., Pending, Proving, Proven).

\item It records the proof for any solved subgoal.

\end{itemize}

This architecture creates a cooperative sprint for each lemma in the plan. When a subgoal becomes active, all Prover agents begin independent tree searches for that single subgoal in parallel. As soon as the first agent finds a valid proof, the subgoal cache records the proof and signals all other agents to terminate their searches, preventing redundant computation. All prover agents then proceed to the next subgoal in the sequence. 

\section{Practical Implementation and Benchmark Results}
\label{sec:prac}
We now present practical implementation of the \texttt{BFS-Prover-V2} system and its benchmark results. 

\textbf{Model and Data:} 
The LLM prover agent is built upon the Qwen2.5-Math-7B and Qwen2.5-32B models~\cite{yang2024qwen2}, which serve as the base for our policy optimization. The multi-stage expert iteration process was initialized with the checkpoint from \texttt{BFS-Prover-V1}~\cite{bfs-prover-v1}. To construct a large-scale training corpus, we autoformalized the NuminaMath-CoT and NuminaMath-1.5 datasets~\cite{numinamath} using carefully designed prompts applied to general-purpose models, augmented with Lean4 compiler feedback. Combined with data provided by Goedel-Prover~\cite{goedel-prover-v1}, this process produced approximately 3 million formal statements. Prompts used for autoformalization can be found in Section~\ref{sec:Automatic_Formalization}. All experiments are conducted in Lean v4.10.0 with LeanDojo~\cite{yang2024leandojo}.

\textbf{Training setup:} We refine the policy LLM after each expert iteration round using one of two Supervised Fine-Tuning (SFT) strategies, chosen based on the outcome of the round. For rounds with a manageable data yield, we perform a continuous finetune from the current best checkpoint for a single epoch, using a conservative cosine learning rate decay from $5 \times 10^{-6}$ and decaying to $1 \times 10^{-7}$. A more comprehensive retrain from the base model is triggered under two conditions: either if the round produces a very large volume of new data, or if model performance has stagnated. In the case of a performance plateau, this retraining is combined with an aggressive data curation step to create a new, refined dataset designed to break the local; see Section~\ref{sec:train}. This retraining process is conducted for $3$ epochs with a higher learning rate decaying from $2 \times 10^{-5}$ and decaying to $1 \times 10^{-6}$. Both strategies utilize a global batch size of 1024.

\textbf{Inference configuration:} Our inference process combines a low-level Prover with a high-level Planner, as detailed in Section~\ref{sec:inference}. The Prover agents utilize a Best-First Search (BFS) algorithm, with an implementation that follows \texttt{BFS-Prover-V1}~\citep{bfs-prover-v1}, where we set the sampling temperature to $1.3$, the expansion width to $3$, and a length normalization factor of $2.0$. For the high-level strategic Planner, we employ Gemini2.5-Pro, while other general-purpose reasoning models can achieve comparable performance if properly prompted. Prompts used for Planner can be found in Section~\ref{sec:Prompts for Planner}.

\textbf{Benchmark results:} We evaluated \texttt{BFS-Prover-V2} on two key benchmarks: miniF2F, a test of high-school competition math, and ProofNet, which challenges reasoning over a large undergraduate-level library. Our system sets a new state of the art for LLM step-provers, achieving 95.08\% on the miniF2F test set (95.49\% on validation) and 41.4\% on the ProofNet test set. The near-saturation performance on miniF2F validates our iterative RL pipeline's ability to master a problem distribution. More importantly, the strong ProofNet result demonstrates successful generalization from the system's training corpus, which consists mainly of high-school competition problems, to the more complex, library-dependent undergraduate problems.
See detailed comparison with other LLM provers in Table~\ref{tab:sota}.

\section{Conclusion}

The primary contributions of this work are the design, implementation, and empirical validation of a holistic system for scaling LLM-based step-provers. On the training side, our multi-stage expert iteration pipeline can overcome common performance plateaus and enable sustained improvement over an extended training period. On the inference side, we introduced a Planner-Prover paradigm that performs subgoal decomposition and parallel tree search. By using a high-level Planner to generate subgoals, we enable the system to tackle complex, multi-step theorems that are intractable for monolithic approaches. The state-of-the-art results on the miniF2F and ProofNet benchmarks provide strong evidence for the efficacy of this integrated approach.

\begin{table}[!htbp]
\begin{center}
\small
\renewcommand{\arraystretch}{1.7}
\fontfamily{bytesansmedium}\selectfont
\begin{tabular}{lcccc}
\Xhline{1.5pt}
Prover Method                            & budget           & miniF2F-test & miniF2F-valid & ProofNet-test \\ \Xhline{1.5pt}
\multicolumn{5}{l}{{\rmfamily\itshape Step-level provers}}                                                            \\ \hline
InternLM2.5-StepProver-7B \cite{intern-prover-v2.5}              & 256 × 32 × 600   & 65.9\%       & 69.6\%        & $\approx$ 27\%    \\ \hline
Hunyuan-Prover-7B \cite{hunyuanprover}               & 600 × 8 × 400    & 68.4\%       & -             & -             \\ \hline
BFS-Prover-V1-7B \cite{bfs-prover-v1}                 & 2048 × 2 × 600   & 70.8\%       & -             & -             \\
                                         & accumulative     & 73.0\%       & -             & -             \\ \hline
MPS-Prover-7B$^\dagger$ \cite{liang2025mps}                           & 64 × 4 × 800 × 8 & 72.54\%      & -             & -             \\
                                         & accumulative     & 75.8\%       & -             & -             \\ \hline
\fontfamily{bytesans}\selectfont BFS-Prover-V2-7B (this work)           & accumulative     & \fontfamily{bytesans}\selectfont 82.4\%      & -             & -             \\ \hdashline
\fontfamily{bytesans}\selectfont BFS-Prover-V2-32B (this work)            & accumulative                & \fontfamily{bytesans}\selectfont 86.1\%       & \fontfamily{bytesans}\selectfont 85.5\%     & \fontfamily{bytesans}\selectfont 41.4\%        \\
\multicolumn{1}{c}{w/ Planner}         & accumulative     & \fontfamily{bytesans}\selectfont 95.1\%       & \fontfamily{bytesans}\selectfont 95.5\%        & -             \\ \Xhline{1.5pt}
\multicolumn{5}{l}{{\rmfamily\itshape Whole-proof provers}}                                                           \\ \hline
DeepSeek-Prover-V2-671B \cite{ds-prover-v2}                  & 8192             & 88.9\%       & 90.6\%        & 37.1\%        \\ \hline
Kimina-Prover-72B$^\dagger$  \cite{kimina-prover}                      & 1024             & 87.7\%       & -             & -             \\
\multicolumn{1}{c}{w/ TTRL search}     & accumulative     & 92.2\%       & -             & -             \\ \hline
Goedel-Prover-32B$^\dagger$ \cite{goedel-prover-v2}                     & 8192             & 92.2\%       & -            & -             \\
\multicolumn{1}{c}{w/ Self-correction} & 1024             & 92.6\%       & -             & -             \\ \hline
Delta-Prover$^\dagger$ \cite{delta-prover}                         & accumulative     & 95.9\%       & -             & -             \\ \hline
Seed-Prover$^\dagger$ \cite{seed-prover}                         & accumulative     & 99.6\%       & -             & -             \\ \Xhline{1.5pt}
\end{tabular}
\caption{Comparison between BFS-Prover-V2 and other leading theorem provers. $^\dagger$ denotes concurrent work.}\label{tab:sota}
\end{center}
\end{table}

\section*{Acknowledgements}
We would like to thank Kai Shen from ByteDance Seed for his insightful discussions throughout this project.

\bibliographystyle{plainnat}
\bibliography{main}

\begin{thebibliography}{43}
\providecommand{\natexlab}[1]{#1}
\providecommand{\url}[1]{\texttt{#1}}
\expandafter\ifx\csname urlstyle\endcsname\relax
  \providecommand{\doi}[1]{doi: #1}\else
  \providecommand{\doi}{doi: \begingroup \urlstyle{rm}\Url}\fi

\bibitem[Anthony et~al.(2017)Anthony, Tian, and Barber]{expert-iteration}
Thomas Anthony, Zheng Tian, and David Barber.
\newblock Thinking fast and slow with deep learning and tree search.
\newblock \emph{Advances in neural information processing systems}, 30, 2017.

\bibitem[Baba et~al.(2025)Baba, Liu, Kurita, and Sannai]{prover-agent}
Kaito Baba, Chaoran Liu, Shuhei Kurita, and Akiyoshi Sannai.
\newblock Prover agent: An agent-based framework for formal mathematical proofs.
\newblock \emph{arXiv preprint arXiv:2506.19923}, 2025.

\bibitem[Bibel et~al.(1993)Bibel, H{\"o}lldobler, and Neugebauer]{history_deduction}
Wolfgang Bibel, Steffen H{\"o}lldobler, and Gerd Neugebauer.
\newblock \emph{Deduction: automated logic}.
\newblock Academic Press London, 1993.

\bibitem[Blokpoel(2024)]{mathlib}
Mark Blokpoel.
\newblock mathlib: A scala package for readable, verifiable and sustainable simulations of formal theory.
\newblock \emph{Journal of Open Source Software}, 9\penalty0 (99):\penalty0 6049, 2024.

\bibitem[Cao et~al.(2025)Cao, Song, Li, Le, Zhang, Xue, and Yang]{dsp+}
Chenrui Cao, Liangcheng Song, Zenan Li, Xinyi Le, Xian Zhang, Hui Xue, and Fan Yang.
\newblock Reviving dsp for advanced theorem proving in the era of reasoning models.
\newblock \emph{arXiv preprint arXiv:2506.11487}, 2025.

\bibitem[Chen et~al.(2025)Chen, Gu, Huang, Huang, Jiang, Jie, Jin, Jin, Li, Ma, et~al.]{seed-prover}
Luoxin Chen, Jinming Gu, Liankai Huang, Wenhao Huang, Zhicheng Jiang, Allan Jie, Xiaoran Jin, Xing Jin, Chenggang Li, Kaijing Ma, et~al.
\newblock Seed-prover: Deep and broad reasoning for automated theorem proving.
\newblock \emph{arXiv preprint arXiv:2507.23726}, 2025.

\bibitem[Comanici et~al.(2025)Comanici, Bieber, Schaekermann, Pasupat, Sachdeva, Dhillon, Blistein, Ram, Zhang, Rosen, et~al.]{gemini2.5}
Gheorghe Comanici, Eric Bieber, Mike Schaekermann, Ice Pasupat, Noveen Sachdeva, Inderjit Dhillon, Marcel Blistein, Ori Ram, Dan Zhang, Evan Rosen, et~al.
\newblock Gemini 2.5: Pushing the frontier with advanced reasoning, multimodality, long context, and next generation agentic capabilities.
\newblock \emph{arXiv preprint arXiv:2507.06261}, 2025.

\bibitem[Geuvers(2009)]{atp_history}
Herman Geuvers.
\newblock Proof assistants: History, ideas and future.
\newblock \emph{Sadhana}, 34\penalty0 (1):\penalty0 3--25, 2009.

\bibitem[Guo et~al.(2025)Guo, Yang, Zhang, Song, Zhang, Xu, Zhu, Ma, Wang, Bi, et~al.]{ds-r1}
Daya Guo, Dejian Yang, Haowei Zhang, Junxiao Song, Ruoyu Zhang, Runxin Xu, Qihao Zhu, Shirong Ma, Peiyi Wang, Xiao Bi, et~al.
\newblock Deepseek-r1: Incentivizing reasoning capability in llms via reinforcement learning.
\newblock \emph{arXiv preprint arXiv:2501.12948}, 2025.

\bibitem[Han et~al.(2021)Han, Rute, Wu, Ayers, and Polu]{han2021proofartifact}
Jesse~Michael Han, Jason Rute, Yuhuai Wu, Edward~W Ayers, and Stanislas Polu.
\newblock Proof artifact co-training for theorem proving with language models.
\newblock \emph{arXiv preprint arXiv:2102.06203}, 2021.

\bibitem[Jiang et~al.(2022)Jiang, Welleck, Zhou, Li, Liu, Jamnik, Lacroix, Wu, and Lample]{dsp}
Albert~Q Jiang, Sean Welleck, Jin~Peng Zhou, Wenda Li, Jiacheng Liu, Mateja Jamnik, Timoth{\'e}e Lacroix, Yuhuai Wu, and Guillaume Lample.
\newblock Draft, sketch, and prove: Guiding formal theorem provers with informal proofs.
\newblock \emph{arXiv preprint arXiv:2210.12283}, 2022.

\bibitem[Lample et~al.(2022)Lample, Lacroix, et~al.]{lample2022hypertree}
Guillaume Lample, Timoth{\'e}e Lacroix, et~al.
\newblock Hypertree proof search for neural theorem proving.
\newblock \emph{Advances in Neural Information Processing Systems}, 35:\penalty0 26337--26349, 2022.

\bibitem[Li et~al.(2024{\natexlab{a}})Li, Beeching, Tunstall, Lipkin, Soletskyi, Huang, Rasul, Yu, Jiang, Shen, et~al.]{numinamath}
Jia Li, Edward Beeching, Lewis Tunstall, Ben Lipkin, Roman Soletskyi, Shengyi Huang, Kashif Rasul, Longhui Yu, Albert~Q Jiang, Ziju Shen, et~al.
\newblock Numinamath: The largest public dataset in ai4maths with 860k pairs of competition math problems and solutions.
\newblock \emph{Hugging Face repository}, 13\penalty0 (9):\penalty0 9, 2024{\natexlab{a}}.

\bibitem[Li et~al.(2024{\natexlab{b}})Li, Du, Song, Li, Wang, Yang, and Mi]{hunyuanprover}
Yang Li, Dong Du, Linfeng Song, Chen Li, Weikang Wang, Tao Yang, and Haitao Mi.
\newblock Hunyuanprover: A scalable data synthesis framework and guided tree search for automated theorem proving.
\newblock \emph{arXiv preprint arXiv:2412.20735}, 2024{\natexlab{b}}.

\bibitem[Liang et~al.(2025)Liang, Song, Li, Yang, Zhang, Mi, and Yu]{liang2025mps}
Zhenwen Liang, Linfeng Song, Yang Li, Tao Yang, Feng Zhang, Haitao Mi, and Dong Yu.
\newblock Mps-prover: Advancing stepwise theorem proving by multi-perspective search and data curation.
\newblock \emph{arXiv preprint arXiv:2505.10962}, 2025.

\bibitem[Lin et~al.(2025{\natexlab{a}})Lin, Tang, Lyu, Wu, Lin, Yang, Li, Xia, Chen, Arora, et~al.]{goedel-prover-v1}
Yong Lin, Shange Tang, Bohan Lyu, Jiayun Wu, Hongzhou Lin, Kaiyu Yang, Jia Li, Mengzhou Xia, Danqi Chen, Sanjeev Arora, et~al.
\newblock Goedel-prover: A frontier model for open-source automated theorem proving.
\newblock \emph{arXiv preprint arXiv:2502.07640}, 2025{\natexlab{a}}.

\bibitem[Lin et~al.(2025{\natexlab{b}})Lin, Tang, Lyu, Yang, Chung, Zhao, Jiang, Geng, Ge, Sun, et~al.]{goedel-prover-v2}
Yong Lin, Shange Tang, Bohan Lyu, Ziran Yang, Jui-Hui Chung, Haoyu Zhao, Lai Jiang, Yihan Geng, Jiawei Ge, Jingruo Sun, et~al.
\newblock Goedel-prover-v2: Scaling formal theorem proving with scaffolded data synthesis and self-correction.
\newblock \emph{arXiv preprint arXiv:2508.03613}, 2025{\natexlab{b}}.

\bibitem[Liu et~al.(2025)Liu, Diao, Lu, Hu, Dong, Choi, Kautz, and Dong]{Prorl}
Mingjie Liu, Shizhe Diao, Ximing Lu, Jian Hu, Xin Dong, Yejin Choi, Jan Kautz, and Yi~Dong.
\newblock Prorl: Prolonged reinforcement learning expands reasoning boundaries in large language models.
\newblock \emph{arXiv preprint arXiv:2505.24864}, 2025.

\bibitem[Moura and Ullrich(2021)]{Lean4}
Leonardo~de Moura and Sebastian Ullrich.
\newblock The lean 4 theorem prover and programming language.
\newblock In \emph{International Conference on Automated Deduction (CADE)}, pages 625--635. Springer, 2021.

\bibitem[OpenAI(2023)]{openai2023gpt4}
OpenAI.
\newblock Gpt-4 technical report.
\newblock \emph{arXiv preprint arXiv:2303.08774}, 2023.

\bibitem[Polu et~al.(2022)Polu, Han, Zheng, et~al.]{polu2022formal}
Stanislas Polu, Jesse~Michael Han, Kunhao Zheng, et~al.
\newblock Formal mathematics statement curriculum learning.
\newblock \emph{arXiv preprint arXiv:2202.01344}, 2022.

\bibitem[Puterman(1990)]{mdp}
Martin~L Puterman.
\newblock Markov decision processes.
\newblock \emph{Handbooks in operations research and management science}, 2:\penalty0 331--434, 1990.

\bibitem[Ren et~al.(2025)Ren, Shao, Song, Xin, Wang, Zhao, Zhang, Fu, Zhu, Yang, et~al.]{ds-prover-v2}
ZZ~Ren, Zhihong Shao, Junxiao Song, Huajian Xin, Haocheng Wang, Wanjia Zhao, Liyue Zhang, Zhe Fu, Qihao Zhu, Dejian Yang, et~al.
\newblock Deepseek-prover-v2: Advancing formal mathematical reasoning via reinforcement learning for subgoal decomposition.
\newblock \emph{arXiv preprint arXiv:2504.21801}, 2025.

\bibitem[Seed et~al.(2025)Seed, Chen, Fan, Liu, Liu, Lin, Wang, Wang, Wei, Xu, et~al.]{seed1.5-thinking}
ByteDance Seed, Jiaze Chen, Tiantian Fan, Xin Liu, Lingjun Liu, Zhiqi Lin, Mingxuan Wang, Chengyi Wang, Xiangpeng Wei, Wenyuan Xu, et~al.
\newblock Seed1. 5-thinking: Advancing superb reasoning models with reinforcement learning.
\newblock \emph{arXiv preprint arXiv:2504.13914}, 2025.

\bibitem[Silver and Sutton(2025)]{era_of_experience}
David Silver and Richard~S Sutton.
\newblock Welcome to the era of experience.
\newblock \emph{Google AI}, 1, 2025.

\bibitem[Silver et~al.(2018)Silver, Hubert, Schrittwieser, Antonoglou, Lai, Guez, Lanctot, Sifre, Kumaran, Graepel, et~al.]{alphazero}
David Silver, Thomas Hubert, Julian Schrittwieser, Ioannis Antonoglou, Matthew Lai, Arthur Guez, Marc Lanctot, Laurent Sifre, Dharshan Kumaran, Thore Graepel, et~al.
\newblock A general reinforcement learning algorithm that masters chess, shogi, and go through self-play.
\newblock \emph{Science}, 362\penalty0 (6419):\penalty0 1140--1144, 2018.

\bibitem[Song et~al.(2024)Song, Yang, and Anandkumar]{lean-copilet}
Peiyang Song, Kaiyu Yang, and Anima Anandkumar.
\newblock Towards large language models as copilots for theorem proving in lean.
\newblock \emph{arXiv preprint arXiv:2404.12534}, 2024.

\bibitem[Sutton(2018)]{sutton2018reinforcement}
Richard~S Sutton.
\newblock Reinforcement learning: An introduction.
\newblock \emph{A Bradford Book}, 2018.

\bibitem[Team et~al.(2025)Team, Du, Gao, Xing, Jiang, Chen, Li, Xiao, Du, Liao, et~al.]{kimi-RL}
Kimi Team, Angang Du, Bofei Gao, Bowei Xing, Changjiu Jiang, Cheng Chen, Cheng Li, Chenjun Xiao, Chenzhuang Du, Chonghua Liao, et~al.
\newblock Kimi k1. 5: Scaling reinforcement learning with llms.
\newblock \emph{arXiv preprint arXiv:2501.12599}, 2025.

\bibitem[Wang et~al.(2025)Wang, Unsal, Lin, Baksys, Liu, Santos, Sung, Vinyes, Ying, Zhu, et~al.]{kimina-prover}
Haiming Wang, Mert Unsal, Xiaohan Lin, Mantas Baksys, Junqi Liu, Marco~Dos Santos, Flood Sung, Marina Vinyes, Zhenzhe Ying, Zekai Zhu, et~al.
\newblock Kimina-prover preview: Towards large formal reasoning models with reinforcement learning.
\newblock \emph{arXiv preprint arXiv:2504.11354}, 2025.

\bibitem[Welleck and Saha(2023)]{llm-step}
Sean Welleck and Rahul Saha.
\newblock Llmstep: Llm proofstep suggestions in lean.
\newblock \emph{arXiv preprint arXiv:2310.18457}, 2023.

\bibitem[Wu et~al.(2024{\natexlab{a}})Wu, Huang, Zhou, Ying, Wang, Lin, and Chen]{intern-prover-v2.5}
Zijian Wu, Suozhi Huang, Zhejian Zhou, Huaiyuan Ying, Jiayu Wang, Dahua Lin, and Kai Chen.
\newblock Internlm2. 5-stepprover: Advancing automated theorem proving via expert iteration on large-scale lean problems.
\newblock \emph{arXiv preprint arXiv:2410.15700}, 2024{\natexlab{a}}.

\bibitem[Wu et~al.(2024{\natexlab{b}})Wu, Wang, Lin, and Chen]{lean-github}
Zijian Wu, Jiayu Wang, Dahua Lin, and Kai Chen.
\newblock Lean-github: Compiling github lean repositories for a versatile lean prover.
\newblock \emph{arXiv preprint arXiv:2407.17227}, 2024{\natexlab{b}}.

\bibitem[Xin et~al.(2024{\natexlab{a}})Xin, Guo, Shao, Ren, Zhu, Liu, Ruan, Li, and Liang]{deepseek_prover_v1}
Huajian Xin, Daya Guo, Zhihong Shao, Zhizhou Ren, Qihao Zhu, Bo~Liu, Chong Ruan, Wenda Li, and Xiaodan Liang.
\newblock Deepseek-prover: Advancing theorem proving in llms through large-scale synthetic data.
\newblock \emph{arXiv preprint arXiv:2405.14333}, 2024{\natexlab{a}}.

\bibitem[Xin et~al.(2024{\natexlab{b}})Xin, Ren, Song, Shao, Zhao, Wang, Liu, Zhang, Lu, Du, et~al.]{deepseek-proer-v1.5}
Huajian Xin, ZZ~Ren, Junxiao Song, Zhihong Shao, Wanjia Zhao, Haocheng Wang, Bo~Liu, Liyue Zhang, Xuan Lu, Qiushi Du, et~al.
\newblock Deepseek-prover-v1. 5: Harnessing proof assistant feedback for reinforcement learning and monte-carlo tree search.
\newblock \emph{arXiv preprint arXiv:2408.08152}, 2024{\natexlab{b}}.

\bibitem[Xin et~al.(2025)Xin, Xi, Yang, Chen, Wu, Xiao, Sun, Zheng, and Shen]{bfs-prover-v1}
Ran Xin, Chenguang Xi, Jie Yang, Feng Chen, Hang Wu, Xia Xiao, Yifan Sun, Shen Zheng, and Kai Shen.
\newblock Bfs-prover: Scalable best-first tree search for llm-based automatic theorem proving.
\newblock \emph{arXiv preprint arXiv:2502.03438}, 2025.

\bibitem[Yang et~al.(2024{\natexlab{a}})Yang, Zhang, Hui, Gao, Yu, Li, Liu, Tu, Zhou, Lin, et~al.]{yang2024qwen2}
An~Yang, Beichen Zhang, Binyuan Hui, Bofei Gao, Bowen Yu, Chengpeng Li, Dayiheng Liu, Jianhong Tu, Jingren Zhou, Junyang Lin, et~al.
\newblock Qwen2. 5-math technical report: Toward mathematical expert model via self-improvement.
\newblock \emph{arXiv preprint arXiv:2409.12122}, 2024{\natexlab{a}}.

\bibitem[Yang et~al.(2024{\natexlab{b}})Yang, Poesia, He, Li, Lauter, Chaudhuri, and Song]{ai4formal}
Kaiyu Yang, Gabriel Poesia, Jingxuan He, Wenda Li, Kristin Lauter, Swarat Chaudhuri, and Dawn Song.
\newblock Formal mathematical reasoning: A new frontier in ai.
\newblock \emph{arXiv preprint arXiv:2412.16075}, 2024{\natexlab{b}}.

\bibitem[Yang et~al.(2024{\natexlab{c}})Yang, Swope, Gu, Chalamala, Song, Yu, Godil, Prenger, and Anandkumar]{yang2024leandojo}
Kaiyu Yang, Aidan Swope, Alex Gu, Rahul Chalamala, Peiyang Song, Shixing Yu, Saad Godil, Ryan~J Prenger, and Animashree Anandkumar.
\newblock Leandojo: Theorem proving with retrieval-augmented language models.
\newblock \emph{Advances in Neural Information Processing Systems}, 36, 2024{\natexlab{c}}.

\bibitem[Ying et~al.(2024)Ying, Wu, Geng, Wang, Lin, and Chen]{lean-workbook}
Huaiyuan Ying, Zijian Wu, Yihan Geng, Jiayu Wang, Dahua Lin, and Kai Chen.
\newblock Lean workbook: A large-scale lean problem set formalized from natural language math problems.
\newblock \emph{arXiv preprint arXiv:2406.03847}, 2024.

\bibitem[Yu et~al.(2025)Yu, Zhang, Zhu, Yuan, Zuo, Yue, Dai, Fan, Liu, Liu, et~al.]{dapo}
Qiying Yu, Zheng Zhang, Ruofei Zhu, Yufeng Yuan, Xiaochen Zuo, Yu~Yue, Weinan Dai, Tiantian Fan, Gaohong Liu, Lingjun Liu, et~al.
\newblock Dapo: An open-source llm reinforcement learning system at scale.
\newblock \emph{arXiv preprint arXiv:2503.14476}, 2025.

\bibitem[Yue et~al.(2025)Yue, Yuan, Yu, Zuo, Zhu, Xu, Chen, Wang, Fan, Du, et~al.]{vapo}
Yu~Yue, Yufeng Yuan, Qiying Yu, Xiaochen Zuo, Ruofei Zhu, Wenyuan Xu, Jiaze Chen, Chengyi Wang, TianTian Fan, Zhengyin Du, et~al.
\newblock Vapo: Efficient and reliable reinforcement learning for advanced reasoning tasks.
\newblock \emph{arXiv preprint arXiv:2504.05118}, 2025.

\bibitem[Zhou et~al.(2025)Zhou, Zhao, Zhang, Wang, Wang, Chen, Wang, Chen, Jie, Zhang, et~al.]{delta-prover}
Yichi Zhou, Jianqiu Zhao, Yongxin Zhang, Bohan Wang, Siran Wang, Luoxin Chen, Jiahui Wang, Haowei Chen, Allan Jie, Xinbo Zhang, et~al.
\newblock Solving formal math problems by decomposition and iterative reflection.
\newblock \emph{arXiv preprint arXiv:2507.15225}, 2025.

\end{thebibliography}

\clearpage

\beginappendix

\section{Case Studies}
\label{sec:cases}

\subsection{Proof Conciseness and Tactic Proficiency}
\label{sec:step_vs_whole}

A primary advantage of our step-level proof approach over the whole-proof paradigm is a dramatic reduction in proof length, which arises from the interactive nature of our method. By engaging with the Lean environment step by step, our model captures and leverages fine-grained tactic state information. This iterative feedback loop significantly improves its ability to employ powerful, high-level tactics such as \texttt{simp}, \texttt{linarith}, and \texttt{nlinarith}, enabling it to solve complex goals in a single step rather than through lengthy, explicit derivations.

To illustrate this contrast, we present a comparative analysis of proofs for two problems, \texttt{aime\_1984\_p7} and \texttt{amc12\_2000\_p1}, generated by \texttt{BFS-Prover-V2} and  \texttt{DeepSeek-Prover-V2}.

\begin{casebox}[frametitle={\texttt{aime\_1984\_p7}}]
For the problem \texttt{aime\_1984\_p7}, \texttt{BFS-Prover-V2} produces a remarkably concise proof, where a single tactic \texttt{simp [h\textsubscript{1}, h\textsubscript{0}]} effectively replaces over 2000 lines that appear in numerous \texttt{have} blocks within \texttt{DeepSeek-Prover-V2}'s proof.

\prooftype{BFS-Prover-V2 (step-level proof):}
\begin{lstlisting}[style=leanstyle, label={lst:aime_step}]
theorem aime_1984_p7
  (f : ℤ → ℤ)
  (h₀ : ∀ n, 1000 ≤ n → f n = n - 3)
  (h₁ : ∀ n, n < 1000 → f n = f (f (n + 5))) :
  f 84 = 997 := by
  simp [h₁, h₀]
\end{lstlisting}

\prooftype{DeepSeek-Prover-V2 (whole-proof):}
\begin{lstlisting}[style=leanstyle, label={lst:aime_whole}]
theorem aime_1984_p7 (f : ℤ → ℤ) (h₀ : ∀ n, 1000 ≤ n → f n = n - 3)
    (h₁ : ∀ n, n < 1000 → f n = f (f (n + 5))) : f 84 = 997 := by
  have h₂ : f 1004 = 1001 := by
    have h₂₁ : f 1004 = 1004 - 3 := by
      apply h₀
      <;> norm_num
    rw [h₂₁]
    <;> norm_num
  
  have h₃ : f 1003 = 1000 := by
    have h₃₁ : f 1003 = 1003 - 3 := by
      apply h₀
      <;> norm_num
    rw [h₃₁]
    <;> norm_num
  
  have h₄ : f 1002 = 999 := by
    have h₄₁ : f 1002 = 1002 - 3 := by
      apply h₀
      <;> norm_num
    rw [h₄₁]
    <;> norm_num
  
  -- ... (approximately 2000 lines of similar calculations omitted) ...
  
  have h₁₉₃ : f 89 = 998 := by
    have h₁₉₃₁ : f 89 = f (f (89 + 5)) := by
      apply h₁
      <;> norm_num
    rw [h₁₉₃₁]
    have h₁₉₃₂ : f (89 + 5) = f 94 := by norm_num
    rw [h₁₉₃₂]
    rw [h₁₉₂]
    <;> norm_num
    <;> linarith
  
  have h₁₉₄ : f 84 = 997 := by
    have h₁₉₄₁ : f 84 = f (f (84 + 5)) := by
      apply h₁
      <;> norm_num
    rw [h₁₉₄₁]
    have h₁₉₄₂ : f (84 + 5) = f 89 := by norm_num
    rw [h₁₉₄₂]
    rw [h₁₉₃]
    <;> norm_num
    <;> linarith
  
  rw [h₁₉₄]
  <;> norm_num
\end{lstlisting}
\end{casebox}

\begin{casebox}[frametitle={\texttt{amc12\_2000\_p1}}]
For the problem \texttt{amc12\_2000\_p1}, \texttt{BFS-Prover-V2} finishes the proof with a concise application of \texttt{nlinarith}, replacing the roughly 80 lines of explicit case analysis over the unknowns \textit{i}, \textit{m}, and \textit{o} required in \texttt{DeepSeek-Prover-V2}'s proof.

\prooftype{BFS-Prover-V2 (step-level proof):}
\begin{lstlisting}[style=leanstyle, label={lst:amc_step}]
theorem amc12_2000_p1
  (i m o : ℕ)
  (h₀ : i ≠ m ∧ m ≠ o ∧ o ≠ i)
  (h₁ : i*m*o = 2001) :
  i+m+o ≤ 671 := by
  obtain ⟨hi, hm, h₂⟩:= h₀
  obtain _ | _ | _ | i := i <;> obtain _ | _ | _ | m := m <;> obtain _ | _ | _ | o := o <;>try omega
  any_goals
    ring_nf at h₁ ⊢; nlinarith;
  ring_nf at * <;> try omega
  nlinarith [mul_nonneg (by positivity : 0 ≤ i) (by positivity : 0 ≤ o), mul_nonneg (by positivity : 0 ≤ m) (by positivity : 0 ≤ i), mul_nonneg (by positivity : 0 ≤ o) (by positivity : 0 ≤ m)]
\end{lstlisting}

\prooftype{DeepSeek-Prover-V2 (whole-proof):}
\begin{lstlisting}[style=leanstyle, label={lst:amc_whole}]
theorem amc12_2000_p1 (i m o : ℕ) (h₀ : i ≠ m ∧ m ≠ o ∧ o ≠ i) (h₁ : i * m * o = 2001) :
    i + m + o ≤ 671 := by

    -- ... (approximately 80 lines of properties related to unknowns i, m, and o omitted) ...
  
  have h_main : i + m + o ≤ 671 := by
    have h₈ : i = 1 ∨ i = 3 ∨ i = 23 ∨ i = 29 ∨ i = 69 ∨ i = 87 ∨ i = 667 ∨ i = 2001 := by
      have h₈₁ : i ∣ 2001 := h₂
      have h₈₂ : i ≤ 2001 := h₅
      have h₈₃ : i > 0 := h_pos_i
      have h₈₄ : i = 1 ∨ i = 3 ∨ i = 23 ∨ i = 29 ∨ i = 69 ∨ i = 87 ∨ i = 667 ∨ i = 2001 := by
        have h₈₅ : i ∣ 2001 := h₈₁
        have h₈₆ : i ≤ 2001 := h₈₂
        interval_cases i <;> norm_num at h₈₅ ⊢ <;> omega
      exact h₈₄
    have h₉ : m = 1 ∨ m = 3 ∨ m = 23 ∨ m = 29 ∨ m = 69 ∨ m = 87 ∨ m = 667 ∨ m = 2001 := by
      have h₉₁ : m ∣ 2001 := h₃
      have h₉₂ : m ≤ 2001 := h₆
      have h₉₃ : m > 0 := h_pos_m
      have h₉₄ : m = 1 ∨ m = 3 ∨ m = 23 ∨ m = 29 ∨ m = 69 ∨ m = 87 ∨ m = 667 ∨ m = 2001 := by
        have h₉₅ : m ∣ 2001 := h₉₁
        have h₉₆ : m ≤ 2001 := h₉₂
        interval_cases m <;> norm_num at h₉₅ ⊢ <;> omega
      exact h₉₄
    have h₁₀ : o = 1 ∨ o = 3 ∨ o = 23 ∨ o = 29 ∨ o = 69 ∨ o = 87 ∨ o = 667 ∨ o = 2001 := by
      have h₁₀₁ : o ∣ 2001 := h₄
      have h₁₀₂ : o ≤ 2001 := h₇
      have h₁₀₃ : o > 0 := h_pos_o
      have h₁₀₄ : o = 1 ∨ o = 3 ∨ o = 23 ∨ o = 29 ∨ o = 69 ∨ o = 87 ∨ o = 667 ∨ o = 2001 := by
        have h₁₀₅ : o ∣ 2001 := h₁₀₁
        have h₁₀₆ : o ≤ 2001 := h₁₀₂
        interval_cases o <;> norm_num at h₁₀₅ ⊢ <;> omega
      exact h₁₀₄
    rcases h₈ with (rfl | rfl | rfl | rfl | rfl | rfl | rfl | rfl) <;>
    rcases h₉ with (rfl | rfl | rfl | rfl | rfl | rfl | rfl | rfl) <;>
    rcases h₁₀ with (rfl | rfl | rfl | rfl | rfl | rfl | rfl | rfl) <;>
    norm_num [mul_assoc, mul_comm, mul_left_comm] at h₁ h₀ ⊢ <;>
    (try omega) <;>
    (try
      {
        norm_num at h₀ ⊢ <;>
        (try omega) <;>
        (try
          {
            ring_nf at h₁ ⊢ <;>
            omega
          })
      }) <;>
    (try
      {
        norm_num at h₀ ⊢ <;>
        (try omega) <;>
        (try
          {
            ring_nf at h₁ ⊢ <;>
            omega
          })
      }) <;>
    (try
      {
        norm_num at h₀ ⊢ <;>
        (try omega) <;>
        (try
          {
            ring_nf at h₁ ⊢ <;>
            omega
          })
      }) <;>
    (try
      {
        norm_num at h₀ ⊢ <;>
        (try omega) <;>
        (try
          {
            ring_nf at h₁ ⊢ <;>
            omega
          })
      })
    <;>
    (try omega)
    <;>
    (try
      {
        ring_nf at h₁ ⊢ <;>
        omega
      })

  exact h_main
\end{lstlisting}
\end{casebox}

\newpage

\subsection{Novel Proof Strategies}
\label{sec:novel_strategies}

Another significant advantage of our step-level proof approach is its ability to discover novel proof strategies that whole-proof or human-proof methods typically would not consider. By exploring the proof space progressively, our system can identify non-obvious connections and construct solutions that are both elegant and insightful. 

We illustrate this capability by examining the problems \texttt{imo\_1963\_p5} and \texttt{algebra\_amgm\_sum1toneqn\_prod1tonleq1}, each of which highlights a distinct advantage of our approach.

\begin{casebox}[frametitle={\texttt{imo\_1963\_p5}}]
For the problem \texttt{imo\_1963\_p5}, our model, DeepSeek-Prover-V2, and Compfiles dataset provide step-level proof, whole-proof, and human-proof versions, respectively. Notably, both whole-proof and human-proof approaches employ similar strategies: multiplying both sides of the equation by $2 \cdot \text{sin}(\pi/7)$, then applying sum-to-product trigonometric identities for simplification. In contrast, \texttt{BFS-Prover-V2} develops an entirely different approach: first transforming the left side of the equation into a polynomial in $\cos(\pi/7)$ using double and triple angle formulas, then proving that $\cos(\pi/7)$ satisfies the corresponding polynomial equation.

\prooftype{BFS-Prover-V2 (step-level proof):}
\begin{lstlisting}[style=leanstyle, label={lst:imo_step}]
theorem imo_1963_p5 :
  Real.cos (π / 7) - Real.cos (2 * π / 7) + Real.cos (3 * π / 7) = 1 / 2 := by
  have x : Real.pi / 7 = Real.pi / 7 * 1 := by ring
  have h : 3 * Real.pi / 7 = Real.pi - 4 * Real.pi / 7 := by ring
  rw [h, cos_sub] <;> norm_num
  have h2 := cos_two_mul (Real.pi / 7)
  have h3 := cos_three_mul (π / 7)
  rw [show 4 * Real.pi / 7 = Real.pi - 3 * Real.pi / 7 by ring,
    cos_sub]
  simp [h2, h3, cos_two_mul, sin_pi, cos_pi]
  ring_nf at h2 h3 ⊢
  norm_num [h2, h3, cos_pi_div_two]
  ring_nf
    <;> have h4 := cos_pi
    <;> simp [h4]
  ring_nf at * <;> norm_num
  rw [← sub_eq_zero]
  nth_rewrite 1 [← sub_eq_zero]
  ring_nf
  apply eq_of_sub_eq_zero
  let y := cos (Real.pi * (1 / 7))
  have:= cos_three_mul (Real.pi * (1 / 7))
  ring_nf at *
  apply eq_of_sub_eq_zero
  clear this h3 h2
  apply eq_of_sub_eq_zero
  have := cos_three_mul (Real.pi * (1 / 7))
  field_simp [mul_assoc] at *
  on_goal 1 => ring
  replace : Real.pi * (1 / 7 : ℝ) = Real.pi / 7 := by ring
  try rw [this]; norm_num
  have h5 := cos_three_mul (Real.pi / 7)
  have : 3 * (Real.pi / 7) = Real.pi - 4 * (Real.pi / 7) := by ring
  simp [this, cos_pi] at h5
  let z := cos (Real.pi / 7)
  rcases lt_trichotomy 0 z with hz | hz | hz
  any_goals simp_all [show cos (4 * (Real.pi / 7)) = cos (2 * (2 * (Real.pi / 7))) by ring,
   cos_two_mul]
  any_goals nlinarith [cos_sq_add_sin_sq (Real.pi / 7), Real.sin_pi_div_two_sub, pow_two_nonneg (cos (Real.pi / 7) - 1),pow_two_nonneg (cos (Real.pi / 7) + 1)]
  contrapose hz
  refine not_lt.2 ?_
  apply cos_nonneg_of_mem_Icc <;> constructor <;> linarith [pi_pos]
\end{lstlisting}

\prooftype{DeepSeek-Prover-V2 (whole-proof):}
\begin{lstlisting}[style=leanstyle, label={lst:imo_whole}]
theorem imo_1963_p5 : Real.cos (Real.pi / 7) - Real.cos (2 * Real.pi / 7) + Real.cos (3 * Real.pi / 7) = 1 / 2 := by
  have h1 : 2 * Real.sin (Real.pi / 7) * (Real.cos (Real.pi / 7) - Real.cos (2 * Real.pi / 7) + Real.cos (3 * Real.pi / 7)) = Real.sin (Real.pi / 7) := by
    have h1₁ : 2 * Real.sin (Real.pi / 7) * Real.cos (Real.pi / 7) = Real.sin (2 * Real.pi / 7) := by
      have h1₁₁ : Real.sin (2 * Real.pi / 7) = 2 * Real.sin (Real.pi / 7) * Real.cos (Real.pi / 7) := by
        have h1₁₂ : Real.sin (2 * Real.pi / 7) = Real.sin (2 * (Real.pi / 7)) := by ring
        rw [h1₁₂]
        have h1₁₃ : Real.sin (2 * (Real.pi / 7)) = 2 * Real.sin (Real.pi / 7) * Real.cos (Real.pi / 7) := by
          rw [Real.sin_two_mul]
          <;> ring
        rw [h1₁₃]
        <;> ring
      linarith
    have h1₂ : 2 * Real.sin (Real.pi / 7) * Real.cos (2 * Real.pi / 7) = Real.sin (3 * Real.pi / 7) - Real.sin (Real.pi / 7) := by
    
  -- ... (approximately 20 lines of calculations omitted) ...
  
    have h1₃ : 2 * Real.sin (Real.pi / 7) * Real.cos (3 * Real.pi / 7) = Real.sin (4 * Real.pi / 7) - Real.sin (2 * Real.pi / 7) := by
    
  -- ... (approximately 20 lines of similar calculations omitted) ...
  
    have h1₄ : Real.sin (4 * Real.pi / 7) = Real.sin (3 * Real.pi / 7) := by
    
  -- ... (approximately 20 lines of similar calculations omitted) ...
  
  have h2 : Real.sin (Real.pi / 7) > 0 := by
    apply Real.sin_pos_of_pos_of_lt_pi
    · linarith [Real.pi_pos, Real.pi_gt_three]
    · linarith [Real.pi_pos, Real.pi_gt_three]
  
  have h3 : Real.cos (Real.pi / 7) - Real.cos (2 * Real.pi / 7) + Real.cos (3 * Real.pi / 7) = 1 / 2 := by
    have h3₁ : 2 * Real.sin (Real.pi / 7) > 0 := by linarith
    have h3₂ : Real.cos (Real.pi / 7) - Real.cos (2 * Real.pi / 7) + Real.cos (3 * Real.pi / 7) = 1 / 2 := by
      apply mul_left_cancel₀ (show (2 * Real.sin (Real.pi / 7) : ℝ) ≠ 0 by linarith)
      nlinarith [Real.sin_le_one (Real.pi / 7), Real.sin_le_one (2 * Real.pi / 7), Real.sin_le_one (3 * Real.pi / 7),
        Real.sin_le_one (4 * Real.pi / 7), Real.sin_le_one (Real.pi / 7)]
    exact h3₂
  
  apply h3
\end{lstlisting}

\prooftype{Compfiles dataset (human-proof):}
\begin{lstlisting}[style=leanstyle, label={lst:imo_human}]
theorem imo1963_p5 :
    Real.cos (π/7) - Real.cos (2*π/7) + Real.cos (3*π/7) = 1/2 := by
  rw [show (2*π/7) = π - (5*π/7) by linarith]
  rw [Real.cos_pi_sub]
  simp only [sub_neg_eq_add]
  have h : 2 * Real.sin (π / 7) ≠ 0 := by
    simp only [ne_eq, mul_eq_zero, OfNat.ofNat_ne_zero, false_or]
    apply ne_of_gt
    apply Real.sin_pos_of_pos_of_lt_pi
    simp only [Nat.ofNat_pos, div_pos_iff_of_pos_right, Real.pi_pos]
    trans 1
    · rw [div_lt_one (by linarith only)]
      linarith only [Real.pi_le_four]
    · linarith only [Real.pi_gt_three]
  apply (mul_right_inj' h).mp
  rw [left_distrib, left_distrib]
  have prod_sum : ∀ (x y : ℝ),
      2 * Real.sin x * Real.cos y = Real.sin (x + y) - Real.sin (y - x) := by
    intro x y
    rw [Real.sin_add, Real.sin_sub]
    linarith only
  rw [prod_sum, prod_sum, prod_sum]
  rw [show (π / 7 + π / 7)     = 2 * π / 7 by linarith only]
  rw [show (π / 7 - π / 7)     = 0         by linarith only]
  rw [show (π / 7 + 5 * π / 7) = 6 * π / 7 by linarith only]
  rw [show (5 * π / 7 - π / 7) = 4 * π / 7 by linarith only]
  rw [show (π / 7 + 3 * π / 7) = 4 * π / 7 by linarith only]
  rw [show (3 * π / 7 - π / 7) = 2 * π / 7 by linarith only]
  rw [Real.sin_zero]
  ring_nf
  rw [← Real.sin_pi_sub]
  rw [show (π - π * (6 / 7)) = π / 7 by linarith]
  congr
  linarith
\end{lstlisting}
\end{casebox}

\begin{casebox}[frametitle={\texttt{algebra\_amgm\_sum1toneqn\_prod1tonleq1}}]
For the problem \texttt{algebra\_amgm\_sum1toneqn\_prod1tonleq1}, the whole-proof model \texttt{DeepSeek-Prover-V2} adopts a standard, first-principles approach: it proceeds by manually handling cases ($n=0$, some $a_i = 0$, all $a_i > 0$), and then takes the logarithm of the product and then applies the well-known inequality $\ln(x) \le x - 1$ to each term, resulting in a verbose proof. In contrast, \texttt{BFS-Prover-V2} recognizes the problem as a special case of the Arithmetic Mean-Geometric Mean (AM-GM) inequality. It directly invokes the corresponding theorem from \texttt{Mathlib}, \texttt{Real.geom\_mean\_le\_arith\_mean}, demonstrating an ability to leverage high-level library theorems for a more insightful and efficient proof.

\prooftype{BFS-Prover-V2 (step-level proof):}
\begin{lstlisting}[style=leanstyle, label={lst:amgm_step}]
theorem algebra_amgm_sum1toneqn_prod1tonleq1
  (a : ℕ → NNReal)
  (n : ℕ)
  (h₀ : ∑ x in Finset.range n, a x = n) :
  ∏ x in Finset.range n, a x ≤ 1 := by
  have g := h₀
  revert h₀
  intro amgm
  let S := Finset.range n
  by_cases h1 : n = 0
  simp[h1]
  have hn  : 0 < n := by omega
  let μ := (fun (x : ℕ) => (a x : ℝ))
  let w : ℕ → ℝ := fun _ => 1
  have w_nonneg : ∀ i, 0 ≤ w i
  simp [w, zero_le_one]
  have w_pos  : 0 < ∏ i in Finset.range n, w i
  all_goals aesop (add simp w)
  have w_nonneg : ∀ x ∈ S, 0 ≤ w x
  any_goals simp [w]
  have w_pos  : 0 < ∏ x ∈ S, w x
  simp [w]
  have amgm_real  : (∏ x ∈ S, μ x ^ (w x : ℝ)) ^ ((∑ x ∈ S, (w x : ℝ))⁻¹) ≤ (∑ x ∈ S, (w x * μ x : ℝ)) / (∑ x ∈ S, (w x : ℝ))
  apply Real.geom_mean_le_arith_mean
  exact w_nonneg
  simp [w, S, hn]
  field_simp[μ]
  simp [μ, w, S] at amgm_real
  norm_cast at amgm_real
  simp[ amgm] at amgm_real
  by_cases H : ∏ a_1 ∈ Finset.range n, a a_1 = 0
  simp [H]
  simp [← NNReal.coe_le_coe, amgm] at amgm_real
  rw  [div_self] at amgm_real
  apply le_of_not_lt
  all_goals aesop (add simp NNReal)
  rw[←NNReal.coe_one] at amgm_real
  norm_cast at amgm_real
  contrapose! amgm_real
  erw [ Real.one_lt_rpow_iff]
  left
  all_goals aesop (add simp [Finset.range])
\end{lstlisting}

\prooftype{DeepSeek-Prover-V2 (whole-proof):}
\begin{lstlisting}[style=leanstyle, label={lst:amgm_whole}]
theorem algebra_amgm_sum1toneqn_prod1tonleq1 (a : ℕ → NNReal) (n : ℕ)
    (h₀ : (∑ x in Finset.range n, a x) = n) : (∏ x in Finset.range n, a x) ≤ 1 := by
  have h₁ : (∏ x in Finset.range n, (a x : ℝ)) ≤ 1 := by
    have h₂ : (∑ x in Finset.range n, (a x : ℝ)) = n := by
      exact_mod_cast h₀
    by_cases h₃ : n = 0
      simp [h₃]
      have h₄ : 0 < n := Nat.pos_of_ne_zero h₃
      by_cases h₅ : ∃ i, i ∈ Finset.range n ∧ (a i : ℝ) = 0
        obtain ⟨i, hi, h₆⟩ := h₅
        have h₇ : (∏ x in Finset.range n, (a x : ℝ)) = 0 := by
          have h₈ : i ∈ Finset.range n := hi
          have h₉ : (a i : ℝ) = 0 := h₆
          have h₁₀ : (∏ x in Finset.range n, (a x : ℝ)) = 0 := by
            calc
              (∏ x in Finset.range n, (a x : ℝ)) = ∏ x in Finset.range n, (a x : ℝ) := rfl
              _ = 0 := by
                apply Finset.prod_eq_zero h₈
                simp [h₉]
          exact h₁₀
        linarith
        have h₅' : ∀ i, i ∈ Finset.range n → (a i : ℝ) > 0 := by
          intro i hi
          have h₆ : ¬∃ i, i ∈ Finset.range n ∧ (a i : ℝ) = 0 := h₅
          have h₇ : (a i : ℝ) ≠ 0 := by
            intro h
            have h₈ : ∃ i, i ∈ Finset.range n ∧ (a i : ℝ) = 0 := ⟨i, hi, by simp_all⟩
            contradiction
          have h₈ : (a i : ℝ) > 0 := by
            have h₉ : (a i : ℝ) ≥ 0 := by exact_mod_cast (a i).prop
            contrapose! h₇
            linarith
          exact h₈
        have h₆ : (∏ x in Finset.range n, (a x : ℝ)) ≤ 1 := by
          have h₇ : ∑ x in Finset.range n, Real.log ((a x : ℝ)) ≤ 0 := by
            have h₈ : ∑ x in Finset.range n, Real.log ((a x : ℝ)) ≤ ∑ x in Finset.range n, ((a x : ℝ) - 1) := by
              apply Finset.sum_le_sum
              intro i hi
              have h₉ : (a i : ℝ) > 0 := h₅' i hi
              have h₁₀ : Real.log ((a i : ℝ)) ≤ (a i : ℝ) - 1 := by
                have h₁₁ : Real.log ((a i : ℝ)) ≤ (a i : ℝ) - 1 := by
                  linarith [Real.log_le_sub_one_of_pos h₉]
                exact h₁₁
              exact h₁₀
              
  -- ... (approximately 40 lines of calculations omitted) ...
  
  have h₂ : (∏ x in Finset.range n, a x) ≤ 1 := by
    have h₃ : (∏ x in Finset.range n, a x : ℝ) ≤ 1 := by
      exact h₁
    have h₄ : (∏ x in Finset.range n, a x : ℝ) = (∏ x in Finset.range n, a x : ℝ) := rfl
    have h₅ : (∏ x in Finset.range n, a x : ℝ) = (∏ x in Finset.range n, (a x : ℝ)) := by simp
    have h₆ : (∏ x in Finset.range n, a x : ℝ) ≤ 1 := by simpa [h₅] using h₁
    have h₇ : (∏ x in Finset.range n, a x : NNReal) ≤ 1 := by
      norm_cast at h₆ ⊢
      <;> simp_all [Finset.prod_range_succ]
      <;> norm_num
      <;> linarith
    simpa using h₇
  exact h₂
\end{lstlisting}
\end{casebox}

Despite the advantages of step-level proof in significantly reducing proof length and discovering novel proof strategies, step-level proof has one notable limitation: poor readability. The interactive nature of step-level proof generation, with its incremental tactic applications and state updates, often results in proofs that are more challenging for humans to follow and understand compared to the more structured and explanatory whole-proof or human-proof approaches. This trade-off between conciseness and readability represents a crucial consideration when evaluating the practical utility of different proof generation paradigms.

\newpage

\section{Illustration of Planner-Prover Paradigm with an IMO Problem}
\label{sec:Planner-Prover}

To demonstrate the effectiveness of our Planner-Prover paradigm, we present an analysis of the solution process for a challenging IMO problem: \texttt{imo\_1969\_p2}.

In the following proof, the statements \texttt{h\_coeffs\_polar}, \texttt{h\_y\_rewritten\_with\_polar}, and \texttt{h\_y\_collapsed\_to\_single\_cos} represent the dynamic replanning phase, while all other \texttt{have} statements belong to the initial planning phase. Unlike in conventional whole-proof methods, \texttt{have} statements in our framework are presented without the \texttt{:= by} clause. This example highlights the crucial role of dynamic replanning in our system. Without dynamic replanning, the prover gets stuck at \texttt{h\_y\_is\_sinusoid}, failing to complete the proof even after 7,200 attempts. With dynamic replanning, however, the system successfully completes the proof in just 800 attempts. The dynamic replanning process breaks down complex steps into smaller, more manageable subgoals, enabling the prover to navigate past critical bottlenecks.

\begin{casebox}[frametitle={\texttt{imo\_1969\_p2} - Part 1}]
\begin{lstlisting}[style=leanstyle]
theorem imo_1969_p2
  (m n : ℝ)
  (k : ℕ)
  (a : ℕ → ℝ)
  (y : ℝ → ℝ)
  (h₀ : 0 < k)
  (h₁ : ∀ x, y x = ∑ i in Finset.range k, ((Real.cos (a i + x)) / (2^i)))
  (h₂ : y m = 0)
  (h₃ : y n = 0) : ∃ t : ℤ, m - n = t * Real.pi := by
  have h_cos_add : ∀ i x, Real.cos (a i + x) = Real.cos (a i) * Real.cos x - Real.sin (a i) * Real.sin x
  simp [cos_add, add_right_inj]

  have h_y_sum_expanded : ∀ (x : ℝ), y x = ∑ i in (Finset.range k : Finset ℕ), (Real.cos (a i) * Real.cos x - Real.sin (a i) * Real.sin x) / ((2 : ℕ) ^ i : ℝ)
  simp [h₁, h_cos_add ]

  have h_y_sum_split : ∀ (x : ℝ), y x = (∑ i in (Finset.range k : Finset ℕ), Real.cos (a i) * Real.cos x / ((2 : ℕ) ^ i : ℝ)) - (∑ i in (Finset.range k : Finset ℕ), Real.sin (a i) * Real.sin x / ((2 : ℕ) ^ i : ℝ))
  intro z <;> simp_rw [h_y_sum_expanded]
  simp [sub_div, Finset.sum_sub_distrib]

  have h_y_expand : ∀ (x : ℝ), y x = (∑ i in (Finset.range k : Finset ℕ), Real.cos (a i) / ((2 : ℕ) ^ i : ℝ)) * Real.cos x - (∑ i in (Finset.range k : Finset ℕ), Real.sin (a i) / ((2 : ℕ) ^ i : ℝ)) * Real.sin x
  intro x_exp
  simp only [Finset.sum_mul, h_y_sum_split]
  congr <;> symm <;> field_simp <;> ring

  have h_k_ge_one : 1 ≤ k
  apply Nat.succ_le_of_lt    <;> exact h₀

  have h_complex_repr : (⟨∑ i in (Finset.range k : Finset ℕ), Real.cos (a i) / ((2 : ℕ) ^ i : ℝ), ∑ i in (Finset.range k : Finset ℕ), Real.sin (a i) / ((2 : ℕ) ^ i : ℝ)⟩ : ℂ) = ∑ i in (Finset.range k : Finset ℕ), Complex.exp (↑(a i) * Complex.I) / ↑(((2 : ℕ) ^ i) : ℝ)
  simp [Complex.exp_mul_I, div_eq_inv_mul, Complex.ext_iff]
  simp [Complex.cos_ofReal_re, Complex.sin_ofReal_re] <;> field_simp <;> norm_cast
  constructor <;> apply Finset.sum_congr <;> aesop
  field_simp [_root_.pow_add, show (4 : ℝ) = 2 ^ 2 by norm_num] <;> ring
  norm_num [mul_comm _ 2, pow_mul]
  rewrite [show (4 : ℝ) ^ x = (2 * 2 : ℝ) ^ x by ring, mul_pow]
    <;> field_simp
    <;> ring

  have h_sum_split : (∑ i in (Finset.range k : Finset ℕ), Complex.exp (↑(a i) * Complex.I) / ↑(((2 : ℕ) ^ i) : ℝ)) = Complex.exp (↑(a 0) * Complex.I) + ∑ i in (Finset.Icc 1 (k-1) : Finset ℕ), Complex.exp (↑(a i) * Complex.I) / ↑(((2 : ℕ) ^ i) : ℝ)
  have h_range_split  : Finset.range k = insert 0 (Finset.Icc 1 (k - 1))
  ext x  <;>  simp  [Nat.lt_succ_iff]
  rcases x with (_|_|x) <;> omega
  rw [h_range_split, Finset.sum_insert]
  norm_num [pow_zero, eq_self_iff_true]
  simp [Nat.le_zero]

  have h_abs_head : Complex.abs (Complex.exp (↑(a 0) * Complex.I)) = 1
  simp [Complex.abs_exp, eq_self_iff_true]

  have h_tail_geom_sum_val : (∑ i in (Finset.Icc 1 (k - 1) : Finset ℕ), 1 / ((2 : ℕ) ^ i : ℝ)) = 1 - 1 / (2 : ℝ) ^ (k - 1)
  have h_tight  : (1 : ℝ) ≤ k
  norm_cast at * <;>
  linarith
  clear h_tight h_sum_split h_complex_repr h_y_expand h_y_sum_split h_y_sum_expanded h_cos_add h₂ h₃ h₁ h₀
  induction' k <;> simp [Finset.sum_Icc_succ_top, *]
  induction' ‹ℕ› <;> simp_all [Finset.sum_Icc_succ_top, pow_succ]
  ring
  <;>ring_nf

  have h_abs_tail_le : Complex.abs (∑ i in (Finset.Icc 1 (k-1) : Finset ℕ), Complex.exp (↑(a i) * Complex.I) / ↑(((2 : ℕ) ^ i) : ℝ)) ≤ 1 - 1 / (2 : ℝ) ^ (k - 1)
  rw [← h_tail_geom_sum_val]
  apply (Complex.abs.sum_le _ _).trans_eq
  apply Finset.sum_congr rfl
  intro i _
  simp [Complex.abs_exp_ofReal_mul_I, Nat.cast_pow, Nat.cast_ofNat]

  have h_abs_tail_lt_one : Complex.abs (∑ i in (Finset.Icc 1 (k-1) : Finset ℕ), Complex.exp (↑(a i) * Complex.I) / ↑(((2 : ℕ) ^ i) : ℝ)) < 1
  refine lt_of_le_of_lt h_abs_tail_le ?_
  refine sub_lt_self _ (by positivity)

  have h_abs_ge_by_rev_triangle : Complex.abs (∑ i in (Finset.range k : Finset ℕ), Complex.exp (↑(a i) * Complex.I) / ↑(((2 : ℕ) ^ i) : ℝ)) ≥ 1 - Complex.abs (∑ i in (Finset.Icc 1 (k-1) : Finset ℕ), Complex.exp (↑(a i) * Complex.I) / ↑(((2 : ℕ) ^ i) : ℝ))
  rw [h_sum_split]
  rw [← h_abs_head]
  apply Complex.abs.le_add
\end{lstlisting}
\end{casebox}

\begin{casebox}[frametitle={\texttt{imo\_1969\_p2} - Part 2}]
\begin{lstlisting}[style=leanstyle]
  have h_abs_ge_final : Complex.abs (∑ i in (Finset.range k : Finset ℕ), Complex.exp (↑(a i) * Complex.I) / ↑(((2 : ℕ) ^ i) : ℝ)) ≥ 1 / (2 : ℝ) ^ (k-1)
  refine' _root_.trans h_abs_ge_by_rev_triangle _
  linarith [h_abs_tail_le]

  have h_abs_gt_zero : 0 < Complex.abs (∑ i in (Finset.range k : Finset ℕ), Complex.exp (↑(a i) * Complex.I) / ↑(((2 : ℕ) ^ i) : ℝ))
  linarith [pow_two_nonneg ((k - 1 : ℕ) : ℝ) ]

  have h_complex_val_ne_zero : (⟨∑ i in (Finset.range k : Finset ℕ), Real.cos (a i) / ((2 : ℕ) ^ i : ℝ), ∑ i in (Finset.range k : Finset ℕ), Real.sin (a i) / ((2 : ℕ) ^ i : ℝ)⟩ : ℂ) ≠ 0
  focus all_goals (norm_num; aesop)
\end{lstlisting}
    
\begin{casebox}[frametitle={}]
\begin{lstlisting}[style=leanstyle]
  have h_coeffs_polar : ∃ (R b : ℝ), 0 < R ∧ (∑ i in (Finset.range k : Finset ℕ), Real.cos (a i) / ((2 : ℕ) ^ i : ℝ)) = R * Real.cos b ∧ (∑ i in (Finset.range k : Finset ℕ), Real.sin (a i) / ((2 : ℕ) ^ i : ℝ)) = R * Real.sin b
  set x := ∑ i ∈ Finset.range k, cos (a i) / ((2 : ℝ) ^ i)
  use Complex.abs (∑ i ∈ Finset.range k, Complex.exp (↑(a i) * Complex.I) / ↑(↑2 ^ i))
  let y : ℝ := ∑ i ∈ Finset.range k, sin (a i) / 2^i
  have h := Complex.abs_mul_cos_add_sin_mul_I (∑ i in Finset.range k, Complex.exp ((a i : ℝ) * Complex.I) / (2 : ℂ) ^ i)
  use  Complex.arg (∑ i in Finset.range k, Complex.exp (↑(a i) * Complex.I) / (2:ℝ) ^ i)
  simp_all [Complex.ext_iff]
\end{lstlisting}
\end{casebox}  

\begin{casebox}[frametitle={}]
\begin{lstlisting}[style=leanstyle]
  have h_y_rewritten_with_polar : ∃ (R a : ℝ), 0 < R ∧ ∀ x, y x = R * Real.cos a * Real.cos x - R * Real.sin a * Real.sin x
  obtain ⟨R, phi, hR_pos, h_cos_eq1, h_sin_eq1⟩ :=  h_coeffs_polar
  use R, phi <;> simp_all[Complex.exp_mul_I, Complex.abs]
\end{lstlisting}
\end{casebox}  

\begin{casebox}[frametitle={}]
\begin{lstlisting}[style=leanstyle]
  have h_y_collapsed_to_single_cos : ∃ (R  a : ℝ), 0 < R ∧ ∀ x, y x = R * Real.cos (x + a) 
  rcases h_y_rewritten_with_polar with ⟨R, a', h_R_pos, h_y_⟩
  use R, a', h_R_pos <;> intros <;> simp [h_y_, cos_add] <;> ring
\end{lstlisting}
\end{casebox} 

\begin{lstlisting}[style=leanstyle]
  have h_y_is_sinusoid : ∃ (R a : ℝ), 0 < R ∧ (∀ x, y x = R * Real.cos (x - a))
  obtain ⟨ R, a, _, hy  ⟩ := h_y_collapsed_to_single_cos
  use R, -a <;> aesop

  have h_roots_exist : ∃ (R a : ℝ), 0 < R ∧ y m = R * Real.cos (m - a) ∧ y n = R * Real.cos (n - a)
  rcases h_y_is_sinusoid with ⟨R, a, h_R_pos, h_y_R_a⟩
  exact ⟨ R, a, h_R_pos,
    by simp [h_y_R_a], by simp [h_y_R_a] ⟩

  have h_cos_zero : ∃ (R a : ℝ), 0 < R ∧ Real.cos (m - a) = 0 ∧ Real.cos (n - a) = 0
  rcases h_roots_exist with ⟨R, a, h_rPos,  h_mEq,  h_nEq⟩
  exact
    ⟨R, a, h_rPos,
      by have := h₂; have := h₃; field_simp [h₁] at * <;> nlinarith,
      by have := h₃; have := h₂; field_simp [h₁] at * <;> nlinarith⟩

  have h_roots_in_pi_half_multiples : ∃ (a : ℝ) (t₁ t₂ : ℤ), m - a = (2 * (t₁ : ℝ) + 1) * Real.pi / 2 ∧ n - a = (2 * (t₂ : ℝ) + 1) * Real.pi / 2
  rcases h_cos_zero with ⟨R, a, _, h_m_cos_zero, h_n_cos_zero⟩
  rw [ cos_eq_zero_iff] at h_m_cos_zero  h_n_cos_zero
  exact ⟨ a,  ↑( Classical.choose h_m_cos_zero ) , ↑( Classical.choose h_n_cos_zero ) , by convert h_m_cos_zero.choose_spec , by convert  h_n_cos_zero.choose_spec ⟩

  have h_m_minus_n_form : ∃ t₁ t₂ : ℤ, m - n = ((2 * (t₁ : ℝ) + 1) * Real.pi / 2) - ((2 * (t₂ : ℝ) + 1) * Real.pi / 2)
  obtain ⟨z, t₁, t₂, h_z_root_m, h_z_root_n⟩ := h_roots_in_pi_half_multiples
  refine ⟨t₁ , t₂,?_⟩<;>
  linarith

  have h_m_minus_n_simplified : ∃ t₁ t₂ : ℤ, m - n = (↑(t₁ - t₂) : ℝ) * Real.pi
  rcases h_m_minus_n_form with ⟨t₁, t₂, h_form⟩  <;>
    exists t₁  <;> exists t₂  <;>  field_simp at h_form ⊢  <;>  linarith
    
  obtain ⟨t₁, t₂,h_m_sub_n_t₁_t₂⟩ := h_m_minus_n_simplified  <;>  use t₁ - t₂  <;>  linarith [h_m_sub_n_t₁_t₂]
\end{lstlisting}
\end{casebox}

\newpage
\section{Prompts Used in This Work}
\label{sec:Prompts}

\subsection{Prompts for Autoformalization}
\label{sec:Automatic_Formalization}

Our autoformalization pipeline operates in two stages to ensure syntactic correctness. First, an \texttt{Initial Formalization Prompt} (shown below) translates a natural language problem into a Lean 4 theorem statement. If the generated code fails to compile, an \texttt{Error Feedback Prompt} is then deployed to revise the statement, using the verbatim error message from the Lean compiler as direct feedback for revision.

\begin{casebox}[frametitle={\texttt{Prompt for Initial Formalization}}]
You are an expert in math proof and the theorem prover: Lean. Given a math problem that contains the question and all conditions, and its corresponding solution that contains solution steps and the correct answer, generate a mathematically equivalent proof problem and rewrite it in the Lean 4 statement. You should follow the following procedures.
\begin{enumerate}
    \item[a):] Identify all questions and conditions in the given problem.
    \item[b):] Identify all solution steps and the correct answers in the given solution.
    \item[c):] With the questions and conditions in a) and correct answers in b), translate the (question, conditions, correct answer) tuple to a mathematically equivalent proof problem that proves question == answer given conditions.
    \item[d):] Rewrite the math proof problem in c) to a Lean 4 statement. Note that you should write the statement only, no proof is required. This also means you do not need to consider the solution steps either.
\end{enumerate}
The first priority is to ensure the generated Lean code can be built successfully. Consider using the following tips.
\begin{itemize}
    \item Use a broader import, e.g., \texttt{import Mathlib}, to bring in the entirety of the necessary library, and remove specific import of submodules, e.g., \texttt{import Mathlib.LinearAlgebra.BasicReal3}, accordingly.
    \item Add \texttt{noncomputable} before \texttt{def} only when necessary.
    \item Use \texttt{by} instead of \texttt{begin end}.
    \item Add \texttt{sorry} to skip the proof.
\end{itemize}
You should strictly follow the below criteria to guarantee the lean statement is equivalent to the mathematical problem.
\begin{itemize}
    \item Each definition used in Lean 4 statement should only directly appear in the conditions problem in a).
    \item Each definition should NOT come from and assume any knowledge directly from the solution step in b).
    \item Each condition in a) should be used as a definition in Lean 4.
    \item For any implications appearing in the conclusions of the original problem, extract their antecedents and declare them as explicit assumptions before the colon, leaving only the consequent in the conclusion after the colon.
    \item For equations, structure the theorem in the form 'conditions : conclusions' where conditions include variable definitions and domains, and conclusions are the solutions to the equation, avoiding implication or equivalence symbols.
\end{itemize}
\noindent\textbf{Below are examples to illustrate the process:}

\noindent\textbf{Example 1 (Number Theory):}\\
\textbf{Lean 4 statement:}
\begin{lstlisting}[style=leanstyle]
theorem nt3_problem (n p : ℕ) (hn : n > 1) (hp : Nat.Prime p)
  (h1 : n ∣ (p - 1)) (h2 : p ∣ (n^6 - 1)) :
  ∃ k : ℕ, (p - n = k^2) ∨ (p + n = k^2) := by
  sorry
\end{lstlisting}
\textbf{problem:}\\
NT3. Let $n>1$ be a positive integer and $p$ a prime number such that $n \mid(p-1)$ and $p \mid(n^{6}-1)$. Prove that at least one of the numbers $p-n$ and $p+n$ is a perfect square.

\noindent\textbf{Example 2 (Number Theory):}\\
\textbf{Lean 4 statement:}
\begin{lstlisting}[style=leanstyle]
theorem nt4_problem (x y : ℕ)
  (hx : x > 0) (hy : y > 0)
  (h1 : ∃ m : ℕ, 3 * x + 4 * y = m^2)
  (h2 : ∃ n : ℕ, 4 * x + 3 * y = n^2) :
  7 ∣ x ∧ 7 ∣ y := by
  sorry
\end{lstlisting}
\textbf{problem:}\\
NT4. If the positive integers $x$ and $y$ are such that both $3x+4y$ and $4x+3y$ are perfect squares, prove that both $x$ and $y$ are multiples of 7.

\noindent\textbf{Example 3 (Algebra):}\\
\textbf{Lean 4 statement:}
\begin{lstlisting}[style=leanstyle]
theorem sum_not_zero (a b c d : ℝ)
  (h1 : a * b * c - d = 1)
  (h2 : b * c * d - a = 2)
  (h3 : c * d * a - b = 3)
  (h4 : d * a * b - c = -6) :
  a + b + c + d ≠ 0 := by
  sorry
\end{lstlisting}
\textbf{problem:}\\
The real numbers $a, b, c, d$ satisfy simultaneously the equations
$a b c-d=1, b c d-a=2, c d a-b=3, d a b-c=-6$.
Prove that $a+b+c+d \neq 0$.

\noindent\textbf{Example 4 (Inequality):}\\
\textbf{Lean 4 statement:}
\begin{lstlisting}[style=leanstyle]
theorem inequality_proof (a b c : ℝ)
  (ha : a > 0) (hb : b > 0) (hc : c > 0) :
  8 / ((a + b)^2 + 4*a*b*c) +
  8 / ((b + c)^2 + 4*a*b*c) +
  8 / ((c + a)^2 + 4*a*b*c) +
  a^2 + b^2 + c^2 ≥
  8 / (a + 3) + 8 / (b + 3) + 8 / (c + 3) := by
  sorry
\end{lstlisting}
\textbf{problem:}\\
The real numbers $a, b, c, d$ satisfy simultaneously the equations
$a b c-d=1, b c d-a=2, c d a-b=3, d a b-c=-6$.
Prove that $a+b+c+d \neq 0$.

\noindent Now, use the same process for the following problem and solution:

\{\textbf{problem}\}

\{\textbf{solution}\}
\end{casebox}

\begin{casebox}[frametitle={\texttt{Prompt for Error Feedback}}]
You are an expert in math proof and the theorem prover: Lean. You are given the following math problem that contains the question and all conditions, and its corresponding solution that contains solution steps and the correct answer.

\{\textbf{problem}\}

\{\textbf{solution}\}

A mathematically equivalent proof problem that proves question == answer given conditions is generated and rewritten in the Lean 4 statement, as shown below:

\{\textbf{Lean 4 statement}\}

However, this lean code got error with \texttt{lake build}, and here is the error message:

\{\textbf{error message}\}

Please modify the lean code to ensure it can be built successfully with \texttt{lake build}. Here is a few tips that might help:
\begin{itemize}
    \item Use a broader import, e.g., \texttt{import Mathlib}, to bring in the entirety of the necessary library, and remove specific import of submodules, e.g., \texttt{import Mathlib.LinearAlgebra.BasicReal3}, accordingly.
    \item Add \texttt{noncomputable} before \texttt{def} only when necessary.
    \item Use \texttt{by} instead of \texttt{begin end}.
    \item Add \texttt{sorry} to skip the proof.
\end{itemize}
\end{casebox}

\subsection{Prompts for Planner}
\label{sec:Prompts for Planner}

\begin{casebox}[frametitle={\texttt{Prompt for Initial Planning}}]
You are an expert assistant specializing in Math Olympiads and the Lean 4 theorem prover.
Your primary goal is to generate \textbf{syntactically perfect, type-checkable} Lean 4 intermediate steps for a given theorem. 
Strictly adhere to the following rules. ANY violation will be considered an error.
While ensuring correctness, generate as many intermediate steps as possible.

\noindent\textbf{Task}\\
  Given the following theorem statement in Lean 4, your job is to \textbf{plan the complete proof} by analyzing the theorem statement and generating a coherent sequence of \texttt{have} statements.
  
  These statements should form a clear chain of reasoning that bridges the theorem’s assumptions to its final claim, breaking down complex arguments into simpler components.

\noindent\textbf{Mandatory Rules}\\
You must comply with every rule in this section. Failure to adhere to any single rule will result in an incorrect output.

\begin{enumerate}
    \item \textbf{Critical Rule: Explicitly Specify Set/Finset Types} \\
    This is the most common and fatal point of error. You must explicitly declare the type for any \texttt{Set} or \texttt{Finset} literal. This rule is non-negotiable.
    \begin{lstlisting}[style=leanstyle, frame=none, backgroundcolor={}, mathescape=true, belowskip=0pt]
- Correct: ({ {-1, 0, 1}} : Set ℤ)
- Incorrect: { {-1, 0, 1}}
    \end{lstlisting}
    \item \textbf{Omit the Proof}: Never provide the proof. Only state the \texttt{have} statement itself.

    \item \textbf{Valid Lean 4 Code}: The entire output block must be type-checkable in a Lean 4.10.0 environment.

    \item \textbf{Use Existing Names}: Use the exact, existing lemma and definition names from \texttt{mathlib}. Do not invent names.

    \item \textbf{No Undeclared Variables}: Do not introduce any variables or constants not declared in the original theorem statement.

    \item \textbf{Explicit Multiplication}: Multiplication must always use the \texttt{*} symbol.
    \begin{lstlisting}[style=leanstyle, frame=none, backgroundcolor={}, mathescape=true, belowskip=0pt]
- Correct: a * x
- Incorrect: ax
    \end{lstlisting}

    \item \textbf{No Chained Inequalities}: Never use chained inequalities. They must be split using logical AND \texttt{$\land$}.
    \begin{lstlisting}[style=leanstyle, frame=none, backgroundcolor={}, mathescape=true, belowskip=0pt]
- Correct: a <= x $\land$ x <= b
- Incorrect: a <= x <= b
    \end{lstlisting}

    \item \textbf{Correct Logarithm Function}: \texttt{Real.log} is only for the natural logarithm. For logarithms with a specified base, you must use \texttt{Real.logb}.
    \begin{lstlisting}[style=leanstyle, frame=none, backgroundcolor={}, mathescape=true, belowskip=0pt]
- Correct: Real.logb (2 : ℝ) 8
- Incorrect: Real.log (2 : ℝ) 8
    \end{lstlisting}

    \item \textbf{Factorial Notation}: In Lean, factorials must be written as \texttt{(n)!} or \texttt{Nat.factorial n}, not \texttt{n!}.
    \begin{lstlisting}[style=leanstyle, frame=none, backgroundcolor={}, mathescape=true, belowskip=0pt]
- Correct: (n)! or Nat.factorial n
- Incorrect: n!
    \end{lstlisting}

    \item \textbf{Numeric Types Must Be Explicitly Annotated}: To avoid type ambiguity in Lean, any expression involving numeric operations must have at least one number's type specified.
    \begin{lstlisting}[style=leanstyle, frame=none, backgroundcolor={}, mathescape=true, belowskip=0pt]
- For division: (1 : ℝ) / 2 = 0.5, but (1 : ℤ) / 2 = 0.
- For subtraction: (1 : ℤ) - 2 = -1, but (1 : ℕ) - 2 = 0.
- Correct: (a : ℝ) / b, a / (b : ℝ), (n : ℤ) - m
- Incorrect: a / b, n - m
    \end{lstlisting}
    
    \item \textbf{Interval Notation}: Do not use \texttt{Icc}, \texttt{Ioo}, \texttt{Ico}, \texttt{Ioc}, etc., to represent intervals. Only use inequalities.
    \begin{lstlisting}[style=leanstyle, frame=none, backgroundcolor={}, mathescape=true, belowskip=0pt]
- Correct: a <= x $\land$ x <= b
- Incorrect: Icc a b
    \end{lstlisting}

    \item \textbf{Complex Numbers}: Use \texttt{Complex.I} for the imaginary unit and \texttt{Complex.abs} for the modulus/absolute value of a complex number.

    \item \textbf{Avoid Common Inequality Theorems}: Avoid using common inequality theorems like Holder's or Jensen's. For inequality problems, try to ensure each proof step only requires basic simplification.

    \item \textbf{Proving Equivalences}: When handling equivalences $\leftrightarrow$ (iff), produce \texttt{have} statements as implications.
\begin{lstlisting}[style=leanstyle, frame=none, backgroundcolor={}, mathescape=true, belowskip=0pt]
- Left-to-right: assume LHS, conclude RHS.
- Right-to-left: assume RHS, conclude LHS.
    \end{lstlisting}

    \item \textbf{Real.pi Notation}: Always use \texttt{Real.pi}, not $\pi$.

    \item \textbf{Final Check}: Before providing the plan, perform a final review to ensure you have scrupulously followed all the rules above, especially the critical rule regarding \texttt{Set}/\texttt{Finset}.
\end{enumerate}

\noindent\textbf{Examples:}\\
Below are examples to illustrate the process and input/output format.

\textbf{Input}:
\begin{lstlisting}[style=leanstyle]
theorem singapore2019_r1_p7 (x : ℝ) (hx : Real.tan x = 5) :
(6 + Real.sin (2 * x)) / (1 + Real.cos (2 * x)) = 83 := by
\end{lstlisting}
\textbf{Output:}
\begin{lstlisting}[style=leanstyle]
have h₁ : Real.sin x = 5 * Real.cos x
have h₂ : Real.sin x ^ 2 = 25 * Real.cos x ^ 2
have h₃ : 26 * Real.cos x ^ 2 = 1
have hsin2x_val : Real.sin (2 * x) = (5 : ℝ) / (13 : ℝ)
have hcos2x_val : Real.cos (2 * x) = -(12 : ℝ) / (13 : ℝ)
\end{lstlisting}

\textbf{Input:}
\begin{lstlisting}[style=leanstyle]
theorem problem4
(g : ℕ → ℝ)
(h : ∀ k : ℕ, 5 ≤ k → k ≤ 124 → g k = (Real.logb (k : ℝ) ((7 : ℝ) ^ (k ^ 2 - 1))) / (Real.logb ((k + 1 : ℝ)) ((7 : ℝ) ^ (k ^ 2 - 4)))) :
(∏ k in Finset.Icc (5 : ℕ) 124, g k) = (41 : ℝ) / 7 := by
\end{lstlisting}
\textbf{Output:}
\begin{lstlisting}[style=leanstyle]
have h_prod_split : (∏ k in (Finset.Icc 5 124 : Finset ℕ), g k) = (∏ k in (Finset.Icc 5 124 : Finset ℕ), ((k ^ 2 - 1) / (k ^ 2 - 4 : ℝ))) * (∏ k in (Finset.Icc 5 124 : Finset ℕ), (Real.logb (k : ℝ) (7 : ℝ) / Real.logb ((k + 1 : ℝ)) (7 : ℝ)))
have h_telescope_part1 : (∏ k in (Finset.Icc 5 124 : Finset ℕ), ((k ^ 2 - 1) / (k ^ 2 - 4 : ℝ))) = (41 : ℝ) / 21
have h_telescope_part2 : (∏ k in (Finset.Icc 5 124 : Finset ℕ), (Real.logb (k : ℝ) (7 : ℝ) / Real.logb ((k + 1 : ℝ)) (7 : ℝ))) = 3
have h_final_product : (41 / 21 : ℝ) * 3 = (41 : ℝ) / 7
\end{lstlisting}

\textbf{Input}:
\begin{lstlisting}[style=leanstyle]
theorem amc12b_variant_p13
(S : Finset ℝ)
(h₀ : ∀ (x : ℝ), x ∈ S ↔ 0 < x ∧ x ≤ 2 * Real.pi ∧ 2 - 4 * Real.sin x + 3 * Real.cos (3 * x) = 0) :
S.card = 4 := by
\end{lstlisting}
\textbf{Output:}
\begin{lstlisting}[style=leanstyle]
have h_interval1 : ∃ x, 0 ≤ x ∧ x < Real.pi / 2 ∧ (2 - 4 * Real.sin x + 3 * Real.cos (3 * x) = 0)
have h_interval2 : ∃ x, Real.pi / 2 ≤ x ∧ x < 3 * Real.pi / 4 ∧ (2 - 4 * Real.sin x + 3 * Real.cos (3 * x) = 0)
have h_interval3 : ∃ x, 3 * Real.pi / 4 ≤ x ∧ x < Real.pi ∧ (2 - 4 * Real.sin x + 3 * Real.cos (3 * x) = 0)
have h_interval4 : ∃ x, Real.pi ≤ x ∧ x < 2 * Real.pi ∧ (2 - 4 * Real.sin x + 3 * Real.cos (3 * x) = 0)
have h_card_eq_4 : S.card = 4
\end{lstlisting}

You must follow all the instructions and mandatory rules above. After deep consideratioin, output the complete plan in Lean for the input below.

\{\textbf{theorem}\}

\end{casebox}

\begin{casebox}[frametitle={\texttt{Prompt for Dynamic Replanning}}]
You are an expert assistant specializing in Math Olympiads and the Lean 4 theorem prover.
Your primary goal is to generate \textbf{syntactically perfect, type-checkable} Lean 4 intermediate steps for a given theorem.
Strictly adhere to the following rules. ANY violation will be considered an error.
While ensuring correctness, generate as many intermediate steps as possible.

\noindent\textbf{Task}\\
Given the following theorem statement, along with already proven subgoals and a currently stuck subgoal, your job is to \textbf{replan the remaining proof} by analyzing the stuck subgoal and generating a coherent sequence of \texttt{have} statements, either by correcting the stuck subgoal or decomposing the stuck subgoal into smaller, logically consistent steps leading toward the theorem's conclusion.

The plan must have all proven subgoals in their original order and position. Only insert new  \texttt{have} statements \textbf{immediately after} them. 

These new statements should form a clear chain of reasoning that bridges the current progress to the final theorem, breaking down complex reasoning into simpler components.

\noindent\textbf{Mandatory Rules}\\
You must comply with every rule in this section. Failure to adhere to any single rule will result in an incorrect output.

(The rules are identical to those in the \texttt{Prompt for Initial Planning}, with the three additional rules shown below introduced before \textbf{Final Check}.)

\begin{enumerate}
    \setcounter{enumi}{15}
    \item \textbf{Insert After Proven Steps}: All new auxiliary \texttt{have} statements must be inserted \textbf{immediately after} the provided proven subgoals. The proven subgoals' wording, order, and placement are \textbf{immutable}. They must remain exactly as given, with no edits, insertions, or extensions inside them.
    
    \item \textbf{Provide Complete Plan}: Output the \textbf{entire updated plan}, preserving the proven subgoals \textbf{exactly as given} and appending new \texttt{have} statements after them in the correct order. Do \textbf{not} output only the new ones. The plan must contain multiple new \texttt{have} statements with sufficient intermediate steps to meaningfully connect to the theorem's conclusion.
    
    \item \textbf{Ensure Logical Continuity}: Each new step must be logically sound. Avoid repetition of the stuck subgoal. If the stuck subgoal is restated, it must not appear immediately after the last proven subgoal. New intermediate steps must be introduced in between.
\end{enumerate}

\textbf{Examples:}

Below are examples to illustrate the process and input/output format.

\textbf{Input}:\\
Theorem
\begin{lstlisting}[style=leanstyle]
theorem singapore2019_r1_p7 (x : ℝ) (hx : Real.tan x = 5) :
(6 + Real.sin (2 * x)) / (1 + Real.cos (2 * x)) = 83 := by
\end{lstlisting}

Proven Subgoals
\begin{lstlisting}[style=leanstyle]
-- (None)
\end{lstlisting}

Stuck Subgoal
\begin{lstlisting}[style=leanstyle, backgroundcolor={}]
have h_sin4x_is_2sin2xcos2x : Real.sin (4 * x) = 2 * Real.sin (2 * x) * Real.cos (2 * x)
\end{lstlisting}

\noindent\textbf{Output:}
\begin{lstlisting}[style=leanstyle]
have h₁ : Real.sin x = 5 * Real.cos x
have h₂ : Real.sin x ^ 2 = 25 * Real.cos x ^ 2
have h₃ : 26 * Real.cos x ^ 2 = 1
have hsin2x_val : Real.sin (2 * x) = (5 : ℝ) / (13 : ℝ)
have hcos2x_val : Real.cos (2 * x) = -(12 : ℝ) / (13 : ℝ)
\end{lstlisting}

\textbf{Input}:\\
Theorem
\begin{lstlisting}[style=leanstyle]
theorem amc12b_variant_p13
(S : Finset ℝ)
(h₀ : ∀ (x : ℝ), x ∈ S ↔ 0 < x ∧ x ≤ 2 * Real.pi ∧ 2 - 4 * Real.sin x + 3 * Real.cos (3 * x) = 0) :
S.card = 4 := by
\end{lstlisting}

Proven Subgoals
\begin{lstlisting}[style=leanstyle]
have h_interval1 : ∃ x, 0 ≤ x ∧ x < Real.pi / 2 ∧ (2 - 4 * Real.sin x + 3 * Real.cos (3 * x) = 0)
have h_interval2 : ∃ x, Real.pi / 2 ≤ x ∧ x < 3 * Real.pi / 4 ∧ (2 - 4 * Real.sin x + 3 * Real.cos (3 * x) = 0)
\end{lstlisting}

Stuck Subgoal
\begin{lstlisting}[style=leanstyle, backgroundcolor={}]
have h_card_eq_4 : S.card = 4
\end{lstlisting}

\noindent\textbf{Output:}
\begin{lstlisting}[style=leanstyle]
have h_interval1 : ∃ x, 0 ≤ x ∧ x < Real.pi / 2 ∧ (2 - 4 * Real.sin x + 3 * Real.cos (3 * x) = 0)
have h_interval2 : ∃ x, Real.pi / 2 ≤ x ∧ x < 3 * Real.pi / 4 ∧ (2 - 4 * Real.sin x + 3 * Real.cos (3 * x) = 0)
have h_interval3 : ∃ x, 3 * Real.pi / 4 ≤ x ∧ x < Real.pi ∧ (2 - 4 * Real.sin x + 3 * Real.cos (3 * x) = 0)
have h_interval4 : ∃ x, Real.pi ≤ x ∧ x < 2 * Real.pi ∧ (2 - 4 * Real.sin x + 3 * Real.cos (3 * x) = 0)
have h_card_eq_4 : S.card = 4
\end{lstlisting}

\textbf{Input}:\\
Theorem:
\begin{lstlisting}[style=leanstyle]
theorem trig_identity_4x (x : ℝ) :
Real.sin (4 * x) = 4 * Real.sin x * Real.cos x * (1 - 2 * Real.sin x ^ 2) := by
\end{lstlisting}

Proven Subgoals:
\begin{lstlisting}[style=leanstyle]
have h_sin4x_is_2sin2xcos2x : Real.sin (4 * x) = 2 * Real.sin (2 * x) * Real.cos (2 * x)
\end{lstlisting}

Stuck Subgoal:
\begin{lstlisting}[style=leanstyle, backgroundcolor={}]
have h_identity : Real.sin (2 * x) * Real.cos (2 * x) = 2 * Real.sin x * Real.cos x * (1 - 2 * Real.sin x ^ 2)
\end{lstlisting}

\noindent\textbf{Output:}
\begin{lstlisting}[style=leanstyle]
have h_sin4x_is_2sin2xcos2x : Real.sin (4 * x) = 2 * Real.sin (2 * x) * Real.cos (2 * x)
have h_sin2x : Real.sin (2 * x) = 2 * Real.sin x * Real.cos x
have h_cos2x_in_terms_of_sin_cos : Real.cos (2 * x) = Real.cos x ^ 2 - Real.sin x ^ 2
have h_cos2x_in_terms_of_sin : Real.cos (2 * x) = 1 - 2 * Real.sin x ^ 2
have h_final_identity : 2 * Real.sin (2 * x) * Real.cos (2 * x) = 4 * Real.sin x * Real.cos x * (1 - 2 * Real.sin x ^ 2)
\end{lstlisting}

You must follow all the instructions and mandatory rules above. After deep consideratioin, output the complete refined plan in Lean for the input below.

\textbf{Input}:\\
Theorem\\
\{\textbf{theorem}\}\\
Proven Subgoals\\
\{\textbf{proven\_subgoals}\}\\
Stuck Subgoal\\
\{\textbf{stuck\_subgoal}\}

\end{casebox}



\end{document}